\documentclass[10pt,twocolumn,letterpaper]{article}

\usepackage[cvpr]{cirl}
\definecolor{cvprblue}{rgb}{0.21,0.49,0.74}
\usepackage[pagebackref,breaklinks,colorlinks,allcolors=cvprblue]{hyperref}
\usepackage{probsurf}
\usepackage[accsupp]{axessibility}

\graphicspath{{figures/}}

\def\paperID{5215} 
\def\confName{CVPR}
\def\confYear{2024}

\title{Objects as volumes: A stochastic geometry view of opaque solids}

\author{Bailey Miller, Hanyu Chen, Alice Lai, Ioannis Gkioulekas\\
Carnegie Mellon University
}

\begin{document}


\maketitle
\begin{abstract}
	We develop a theory for the representation of opaque solids as volumes. Starting from a stochastic representation of opaque solids as random indicator functions, we prove the conditions under which such solids can be modeled using exponential volumetric transport. We also derive expressions for the volumetric attenuation coefficient as a functional of the probability distributions of the underlying indicator functions. We generalize our theory to account for isotropic and anisotropic scattering at different parts of the solid, and for representations of opaque solids as stochastic implicit surfaces. We derive our volumetric representation from first principles, which ensures that it satisfies physical constraints such as reciprocity and reversibility. We use our theory to explain, compare, and correct previous volumetric representations, as well as propose meaningful extensions that lead to improved performance in 3D reconstruction tasks.
\end{abstract}
\vspace{-2em}
\section{Introduction}\label{sec:intro}
\vspace{-0.5em}

Volumetric representations have a long history in applied physics \citep{mishchenko2006multiple} and computer graphics \citep{novak2018monte}, where they enable efficient light transport simulation in translucent objects (e.g., tissue, clouds, materials such as wax and soap) and participating media (e.g., smoke, fog, murky water). Since the introduction of NeRF by \citet{mildenhall2021nerf}, there has been a proliferation of neural rendering techniques~\citep{wang2021neus, yariv2021volume, oechsle2021unisurf, Azinovic_2022_CVPR, fu2022geoneus, Rematas2021UrbanRF, yu2022monosdf, wang2022neuris} that use volumetric representations for scenes much unlike the above examples, comprising opaque objects (rather than translucent ones) in free space (rather than volumetric media). The tremendous success of volumetric representations for scenes without subsurface or volumetric scattering motivates questions such as: Why can we use volumetric light transport to simulate scenes with only light-surface interactions? What is the mathematical underpinning for modeling an opaque object as a volume? What are the properties of such a volume? 

Our goal is to answer these questions. We start from first principles, revisiting the derivation of volumetric representations for translucent objects and participating media. As recent work in computer graphics highlights~\citep{bitterli18framework,jarabo2018radiative,d2018reciprocal}, volumetric representations are a formalism for querying \emph{stochastic geometry}~\citep{preisendorfer2014radiative,chiu2013stochastic}: From this lens, 
volumetric quantities such as transmittance and free-flight distribution are the answers to queries such as ``are two points mutually visible'' (a visibility query) and ``what is the distance to first intersection along a ray'' (a ray-casting query), when the geometry occluding visibility and terminating rays is stochastic.

Volumetric representations for translucent and participating media are stochastic abstractions of their microscopic structure: Such media comprise numerous microscopic particles that reflect and occlude light rays. Modeling explicit microparticle configurations, and rendering light transport through them, is prohibitively expensive. As a compromise for efficiency, volumetric representations allow to simulate light transport in such media \emph{on average}~\citep{bar2019monte}. These representations replace explicit with \emph{statistical} descriptions of microparticle configurations (e.g., average particle location, size, shape, and orientation), analogously to statistical BRDF models for surface microgeometry~\citep{cook1982reflectance,oren1993diffuse,walter2007microfacet,heitz2016multiple,dupuy2016}. Computer graphics has developed volumetric representations for microparticle media that account for details such as varying particle size and material~\citep{frisvad2007computing,levis2017multiple}, shape and orientation~\citep{jakob2010anisotropy,heitz2015sggx}, and placement correlations~\citep{bitterli18framework,jarabo2018radiative,d2018reciprocal}.

We develop (\cref{sec:solids}) analogous volumetric representations for scenes comprising opaque macroscopic 3D objects, or \emph{opaque solids}, using stochastic geometry theory. 
We prove (\cref{sec:exponential}) formal conditions for \emph{exponential} volumetric representations to apply to stochastic opaque solids; and functional relationships between volumetric parameters (i.e., attenuation coefficient) and stochastic geometry models. We adapt (\cref{sec:anisotropy}) anisotropic volumetric representations of microparticle geometry to opaque solids, to account for effects such as directionally-dependent foreshortening near surfaces, and directionally-independent attenuation far from them. We extend (\cref{sec:implicit}) our volumetric representations to utilize geometry models common in current practice (e.g., implicit surfaces). Our theory delivers volumetric representations equipped with properties necessary for physical plausibility (e.g., reciprocity and reversibility).

Our work is not the first to derive volumetric representations for opaque solids~\citep{wang2021neus,yariv2021volume,oechsle2021unisurf,sellan2022stochastic}. Previous derivations generally consider how to transform a deterministic geometry representation (e.g., signed distance function) into a volumetric one that behaves approximately like the deterministic geometry. Despite its empirical success, this methodological approach remains \emph{heuristic} and requires arbitrary choices (e.g., deciding what properties of deterministic geometry to preserve in the volumetric representation).
By contrast, our derivation is \emph{rigorous}, starting from only the axioms of volumetric transport, and helps place this prior work on a solid mathematical footing: We show (\cref{sec:prior}) that our theory explains previous volumetric representations as special cases of ours, corresponding to different stochastic modeling choices for the underlying opaque geometry. Our theory additionally highlights and addresses critical defects of previous volumetric representations (e.g., lack of reciprocity and reversibility), and generalizes them in principled ways. 

Our volumetric representation can be readily incorporated within existing volumetric neural rendering pipelines \citep{wang2021neus,nerfstudio,Yu2022SDFStudio,multinerf2022}. 
We show experimentally (\cref{sec:experiments}) that replacing previous volumetric representations \citep{yariv2021volume,wang2021neus} with ours leads to significantly better (qualitatively and quantitatively) 3D reconstructions on common datasets. We provide interactive visualizations, open-source code, and a supplement with all appendices on the project website.%
\footnote{{\scriptsize \url{https://imaging.cs.cmu.edu/volumetric_opaque_solids}}}



\vspace{-0.75em}
\section{Volumetric light transport background}\label{sec:rte}
\vspace{-0.5em}

We begin with background on \emph{volumetric light transport} (also known as radiative transfer). We follow \citet{bitterli18framework} for our review, and refer to \citet[Chapter XV]{preisendorfer2014radiative} and \citet{chiu2013stochastic} for a more comprehensive discussion.

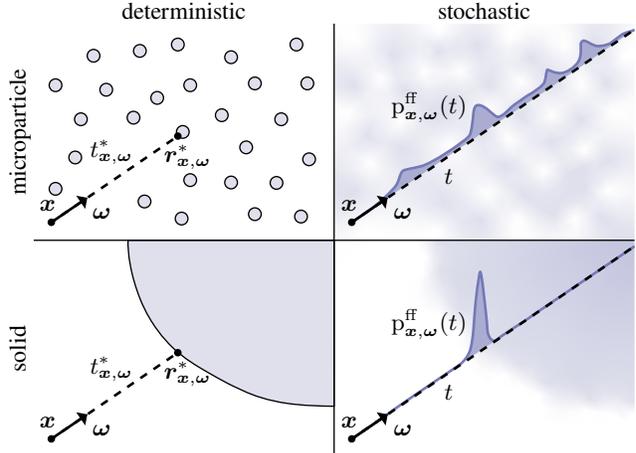
\begin{figure}[t]
	\centering
	\includegraphics[width=\columnwidth]{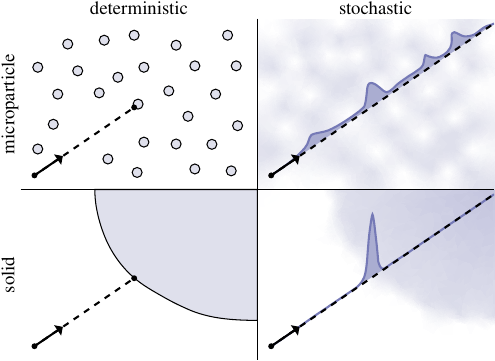}
	\put(-22.5em,9.3em){{\small $\point$}}
	\put(-22.5em,9.3em-82pt){{\small $\point$}}
	\put(-22.5em+114pt,9.3em){{\small $\point$}}
	\put(-22.5em+114pt,9.3em-82pt){{\small $\point$}}
	\put(-20.5em,9em){{\small $\direction$}}
	\put(-20.5em,9em-82pt){{\small $\direction$}}
	\put(-20.5em+114pt,9em){{\small $\direction$}}
	\put(-20.5em+114pt,9em-82pt){{\small $\direction$}}
	\put(-17.8em,11.15em){{\small $\intersection_{\point,\direction}$}}
	\put(-17.8em,11.15em-82pt){{\small $\intersection_{\point,\direction}$}}
	\put(-20.6em,11.35em){{\small $\casting_{\point,\direction}$}}
	\put(-20.6em,11.35em-82pt){{\small $\casting_{\point,\direction}$}}
	\put(-20.6em+114pt,13em){{\small $\freeflight_{\point,\direction}\paren{\distance}$}}
	\put(-20.6em+114pt,13em-82pt){{\small $\freeflight_{\point,\direction}\paren{\distance}$}}
	\put(-18.6em+114pt,10.35em){{\small $\distance$}}
	\put(-18.6em+114pt,10.35em-82pt){{\small $\distance$}}
	\vspace{-1em}
	\caption{Volumetric representations replace deterministic (left) with stochastic (right) ray casting: rather than find the first intersection with deterministic geometry, they use the free-flight distribution along a ray to represent the probability of first intersection with stochastic geometry. Classical volumetric representations describe stochastic microparticle geometry (top). We derive volumetric representations for stochastic solid geometry (bottom).}
	\label{fig:background}
	\vspace{-1.5em}
\end{figure}

\paragraph{Setup.} Volumetric light transport models scenes with \emph{stochastic} geometry (\cref{fig:background}). Classically in computer graphics, these are scenes comprising numerous microscopic particles~\citep{moon2007rendering} (e.g., translucent materials); whereas in neural rendering, they are scenes comprising macroscopic opaque objects. We term the two settings \emph{stochastic microparticle geometry} and \emph{stochastic solid geometry}, respectively.

In both settings, volumetric light transport algorithms simulate \emph{expected} radiometric measurements over all realizations of the stochastic geometry~\citep{bar2019monte}.
They leverage the fact that deterministic light transport algorithms (e.g., path tracing) interact with scene geometry only through two geometric queries:
\begin{enumerate*}[label=\textbf{Q\arabic*.},ref={Q\arabic*}]
	\item \label{enum:vis} \emph{visibility queries} to compute the \emph{visibility} $\visibility\paren{\point,\pointalt} \in \curly{0,1}$ between points $\point, \pointalt \in \R^3$;
	\item \label{enum:ray} \emph{ray-casting queries} to compute the \emph{free-flight distance} $\casting_{\point, \direction} \in \leftinc{0, \infty}$ a ray with origin $\point\in\R^3$ and direction $\direction \in \Sph^2$ travels until it first intersects the scene.
\end{enumerate*}
Thus, we can translate light transport algorithms from the deterministic to the volumetric setting by \emph{stochasticizing} these geometric queries \citep{bitterli18framework}. 

To facilitate discussion of these stochastic queries, we introduce some notation. We denote by 
$\ray_{\point, \direction}\paren{\distance} \equiv \point + \distance \cdot \direction$
the point on a ray with origin $\point \in \R^3$ and direction $\direction \in\Sph^2$ after travel distance $\distance\in\leftinc{0, \infty}$; and by $\visibility_{\point, \direction}\paren{\distance}\equiv \visibility\paren{\point, \ray_{\point, \direction}\paren{\distance}}$ the visibility along the ray. Then, the free-flight distance becomes $\casting_{\point,\direction} \equiv \max\curly{\distance\in\leftinc{0,\infty}: \visibility_{\point, \direction}\paren{\distance} = 1}$, and we denote by $\intersection_{\point, \direction} \equiv \ray_{\point, \direction}\paren{\casting_{\point, \direction}}$ the first intersection point. 
%
\begin{definition}\label{def:general_defs}
    In a scene with stochastic geometry $\geometry$, the \emph{transmittance} along a ray $\ray_{\point, \direction}\paren{\distance}$ is the probability of visibility from the ray origin $\point$---equivalently, the tail distribution of the free-flight distance $\casting_{\point, \direction}$:
	%
	\begin{equation}\label{eqn:transmittance_ray}
		\!\transmittance_{\point, \direction}\paren{\distance} \equiv \Prob_{\geometry}\curly{\visibility_{\point, \direction}\paren{\distance} = 1} = \Prob_{\geometry}\curly{\casting_{\point, \direction} \ge \distance}.
	\end{equation}
	The \emph{free-flight distribution} along a ray is the probability density function of the free-flight distance $\casting_{\point, \direction}$:
	\begin{equation}\label{eqn:free_flight}
		\freeflight_{\point, \direction}\paren{\distance} \equiv - \frac{\ud \transmittance_{\point, \direction}}{\ud \distance}\paren{\distance}.
	\end{equation}
	The \emph{attenuation coefficient} at point $\point$ and direction $\direction$ is the probability density of zero free-flight distance (equivalently, probability density of ray termination through $\point$ along $\direction$):
	\begin{equation}
		\coeff\paren{\point, \direction} \equiv \freeflight_{\point, \direction}\paren{0}.\label{eqn:coeff}
	\end{equation}
\end{definition}
\vspace{-10pt}
We postpone definitions of stochastic geometry $\geometry$ till \cref{sec:solids}. The transmittance inherits the following properties from visibility:
\begin{enumerate*}
	\item it is \emph{reciprocal}, $\transmittance_{\point,\direction}\paren{t} = \transmittance_{\pointalt,-\direction}\paren{t}$ if $\pointalt \equiv \ray_{\point,\direction}\paren{\distance}$;
	\item it is monotonically non-increasing, $\transmittance_{\point,\direction}\paren{t} \le \transmittance_{\point,\direction}\paren{s}$ if $t < s$; 
	\item it satisfies $\transmittance_{\point,\direction}\paren{0} = 1$.
\end{enumerate*}
The transmittance and free-flight distribution generalize the visibility (\labelcref{enum:vis}) and ray-casting (\labelcref{enum:ray}) queries, respectively: for deterministic geometry, \cref{eqn:transmittance_ray}  reduces to the deterministic visibility, and \cref{eqn:free_flight} reduces to the Dirac delta distribution $\delta\paren{\distance - \casting_{\point, \direction}}$ centered at the deterministic free-flight distance. The attenuation coefficient will become important when we discuss exponential transport below.

\begin{figure*}[t]
	\centering
	\includegraphics[width=\textwidth]{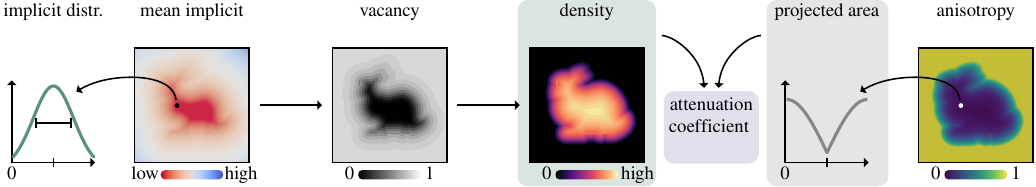}
	\put(-463.7212pt,55.4619pt){{\small Eq. \labelcref{eqn:specialize0}}}
	\put(-86.9085pt,55.4619pt){{\small Eq. \labelcref{eqn:projected_mixture}}}
	\put(-372pt,43pt){{\small Eq. \labelcref{eqn:vacancy_implicit}}}
	\put(-277.1245pt,43pt){{\small Eq. \labelcref{eqn:coeff_implicit}}}
	\put(-170.1365pt,75.1221pt){{\small Eq. \labelcref{eqn:coeff_generalized}}}
	\put(-118pt,72pt){{\small $\projected_{\ndf}\paren{\point, \direction}$}}
	\put(-37pt,72pt){{\small $\anisotropy\paren{\point}$}}
	\put(-226pt,72pt){{\small $\density\paren{\point}$}}
	\put(-318pt,72pt){{\small $\vacancy\paren{\point}$}}
	\put(-413pt,72pt){{\small $\meanimplicit\paren{\point}$}}
	\put(-492pt,72pt){{\small $\pdf_{\implicit\paren{\point}}\paren{\imval}$}}
	\put(-170.1365pt,16pt){{\small $\coeff\paren{\point,\direction}$}}
	\put(-479.5pt,3.5pt){{\small $\meanimplicit\paren{\point}$}}
	\put(-473.5pt,33pt){{\small $s$}}
	\put(-454pt,3.5pt){{\small $\imval$}}
	\put(-85pt,3.5pt){{\small $\pi$}}
	\put(-105pt,3.5pt){{\small $\nicefrac{\pi}{2}$}}
	\put(-420pt,33pt){{\small $\point$}}
	\put(-43.7469pt,33pt){{\small $\textcolor{white}{\point}$}}
	\vspace{-0.5em}
	\caption{Overview of our theory, presented in \cref{thm:coefficient,def:anisotropy,pro:implicit}.}
	\vspace{-1.5em}
	\label{fig:theory}
\end{figure*}

We can use these definitions to generalize deterministic light transport algorithms, which recursively use the equation for the conservation of radiance along a ray, $L_\mathrm{i}\paren{\point, \direction} = L_\mathrm{o}\paren{\intersection_{\point, \direction}, -\direction}$,
%
%
to volumetric light transport algorithms, which recursively use the \emph{expectation} of this equation:
\begin{align}
	&\Exp{\geometry}{L_\mathrm{i}\paren{\point, \direction}} = \Exp{\geometry}{L_\mathrm{o}\paren{\intersection_{\point, \direction}, -\direction}} \\
	&\quad = \int_0^\infty \mymathbox{intBlueL}{intBlueLL}{geometry}{\freeflight_{\point, \direction}\paren{\distance}}\cdot\mymathbox{intCrimL}{intCrimLL}{global illumination}{\vphantom{\freeflight_{\point, \direction}}\Exp{\geometry}{\conditional{L_\mathrm{o}\paren{\ray_{\point, \direction}\paren{\distance}, -\direction}}{\distance}}} \ud \distance.\label{eqn:volume}
\end{align}
If we drop the distinction between expected and actual radiances, \cref{eqn:volume} is the \emph{volume rendering equation} that neural volume rendering techniques use~\citep{mildenhall2021nerf}. Such techniques typically separately model the geometry and global illumination terms, the latter as either volumetric emission \citep{mildenhall2021nerf} or in-scattered radiance \citep{verbin2022ref}. We focus on the geometry term, but discuss in \cref{sec:illumination} implications for the global illumination term from geometry modeling choices. 
%

\paragraph{Exponential transport.} Most commonly in computer vision and graphics, the free-flight distance is an \emph{exponential random variable}, an assumption we call \emph{exponential transport}. Then, \crefrange{eqn:transmittance_ray}{eqn:coeff} and transmittance reciprocity imply:
\vspace{-10pt}
\begin{align}
    \transmittance_{\point, \direction}\paren{\distance} &= \exp\paren{-\int_0^\distance \coeff\paren{\ray_{\point,\direction}\paren{\distanceVariable}, \direction} \ud \distanceVariable}, \label{eqn:transmittance_exponential} \\
    \freeflight_{\point, \direction}\paren{\distance} &= \coeff\paren{\ray_{\point,\direction}\paren{\distance}, \direction} \transmittance_{\point, \direction}\paren{\distance}, \label{eqn:freeflight_exponential} \\
    \coeff\paren{\point, \direction} &= \coeff\paren{\point, -\direction}. \label{eqn:reciprocity_sigma}
\end{align}
Thus, the attenuation coefficient becomes the \emph{rate parameter} of the 
free-flight distance. Given known coefficient values, there exist efficient and accurate numerical approximations for the free-flight distribution and transmittance~\citep{georgiev2019integral,kettunen2021unbiased,novak2014residual}. 

Exponential transport has been extensively studied for stochastic microparticle geometry~\citep{mishchenko2006multiple}: It is equivalent to the \emph{Poisson Boolean model} of stochastic geometry, where microparticle locations are independent~\citep{jarabo2018radiative,bitterli18framework,d2018reciprocal} and distributed as a spatial Poisson process \citep{chiu2013stochastic,last2017lectures}. This model allows expressing the attenuation coefficient analytically as a function of the probability distributions for the particle location, size, material, shape, and orientation~\citep{frisvad2007computing,jakob2010anisotropy,heitz2015sggx}. The recent success of exponential transport in neural rendering~\citep{mildenhall2021nerf,wang2021neus,yariv2021volume,oechsle2021unisurf} motivates our study in \cref{sec:solids}, where we derive, for the first time, exponential transport models for stochastic solid geometry.
Notably, \citet{vicini2021non} suggest using non-exponential transport for stochastic solid geometry, a suggestion we briefly discuss in \cref{sec:conclusion}.

\paragraph{Isotropic and anisotropic transport.} In isotropic transport, the attenuation coefficient 
is independent of direction, $\coeff\paren{\point, \direction} = \coeff\paren{\point}$; and conversely for anisotropic transport~\citep{jakob2010anisotropy}. In stochastic microparticle geometry, isotropic transport models microparticles as rotationally-symmetric scatterers (spheres). We explain isotropic versus anisotropic transport for stochastic solid geometry in \cref{sec:anisotropy}.

\vspace{-0.75em}
\section{Stochastic opaque solids}\label{sec:solids}
\vspace{-0.5em}

We develop our exponential transport theory for stochastic solid geometry in three parts:
\begin{enumerate*}
	\item In \cref{sec:exponential}, we introduce a stochastic model for solid geometry, prove conditions for exponential transport, and derive expressions for the attenuation coefficient. 
	\item In \cref{sec:anisotropy}, we generalize these expressions to model variable anisotropic behavior. 
	\item In \cref{sec:implicit}, we adapt our expressions to implicit-surface geometry representations.
\end{enumerate*}
\Cref{fig:theory} summarizes our theory, and the project website includes a video explanation.

\vspace{-0.5em}
\subsection{Conditions for exponential transport} \label{sec:exponential}
\vspace{-0.5em}

To formalize our exponential transport model for stochastic \emph{opaque} solid geometry, we first define an opaque solid.%
\footnote{We borrow the term \emph{solid} from \citet{koenderink1990solid}.}
\begin{definition}\label{def:opaque_solid}
	We define an \emph{indicator function} $\indicator : \R^3 \to \curly{0,1}$ as a binary scalar field, and associate with it a \emph{solid} $\solid \equiv \curly{\point \in \R^3 : \indicator\paren{\point} = 1}$. The solid $\solid$ is \emph{opaque} if and only if, for every point $\point \in \solid$ and direction $\direction \in \Sph^2$, the visibility satisfies $\visibility_{\point, \direction}\paren{\distance} = \delta\paren{\distance}$.
	\vspace{-5pt}
\end{definition}
The definition of opacity implies that no ray of light can reach points inside the solid $\solid$. Therefore, our volumetric light transport formulation will exclude refractive surfaces. 
We can now use Definition~\ref{def:opaque_solid} to also define a \emph{stochastic solid}.

\begin{definition}\label{def:stochastic_solid}
	When the indicator function $\indicator\paren{\point}$ is a \emph{random} scalar field, we call the associated solid $\solid$ a \emph{stochastic solid}, for which we define the \emph{occupancy} $\occupancy : \R^3 \to \bracket{0,1}$ and \emph{vacancy} $\vacancy : \R^3 \to \bracket{0,1}$ as the scalar fields:
	\begin{align}
		\occupancy\paren{\point} &\equiv \Prob\curly{\indicator\paren{\point} = 1}, \label{eqn:occupancy} \\
		\vacancy\paren{\point} &\equiv \Prob\curly{\indicator\paren{\point} = 0} = 1 - \occupancy\paren{\point}.\label{eqn:vacancy}
	\end{align}
	\vspace{-20pt}
\end{definition}
With this definition, we can interpret probabilities involving $\solid$ in \crefrange{eqn:transmittance_ray}{eqn:coeff} as probabilities over all realizations of the random indicator function $\indicator$. We will consider the restriction of the indicator function, occupancy, and vacancy on a ray with origin $x \in \R^3$ and direction $\direction \in \Sph^2$: $\indicator_{\point, \direction}\paren{\distance} \equiv \indicator\paren{\ray_{\point, \direction}\paren{\distance}}$, and analogously for $\occupancy_{\point, \direction}\paren{\distance}$ and $\vacancy_{\point, \direction}\paren{\distance}$. We can now state our main technical result.

\begin{thm}[label={the:coefficient}]{Exponential transport in opaque solids}{coefficient}
	We assume a random indicator function $\indicator$ and associated stochastic opaque solid $\solid$. Then, for any ray with origin $\point \in \R^3$ and direction $\direction \in \Sph^2$, the free-flight distribution $\freeflight_{\point, \direction}$ is exponential if and only if the restriction of the indicator function on this ray, $\indicator_{\point, \direction}$, is a \emph{continuous-time discrete-space Markov process}; that is, it satisfies:
	\begin{align}\label{eqn:markov}
		\Prob\curly{\conditional{\indicator_{\point, \direction}\paren{\distance}}{\indicator_{\point, \direction}\paren{\distance_n}, \distance_n < \distance, n=1,\dots,N}}& \nonumber \\
		= \Prob\curly{\conditional{\indicator_{\point, \direction}\paren{\distance}}{\indicator_{\point, \direction}\paren{\max\displaystyle_n\distance_n}}}&.
	\end{align}
	Additionally, the process $\indicator_{\point, \direction}$ is reversible and the corresponding transmittance $\transmittance_{\point, \direction}$ is reciprocal if and only if the attenuation coefficient $\coeff$ equals:
	\vspace{-5pt}
	\begin{equation}\label{eqn:coeff_markov}
		\mymathboxnt{intBlueLL}{\coeff_\delta\paren{\point, \direction} \equiv \abs{\direction \cdot \nabla \log\paren{\vacancy\paren{\point}}}\! =\! \frac{\abs{\direction \cdot \nabla \vacancy\paren{\point}}}{\vacancy\paren{\point}}}.
	\end{equation}
\end{thm}
\vspace{-0.25em}
We explain the notation $\coeff_\delta$ in \cref{sec:anisotropy}. Expressions for transmittance and free-flight distribution follow from \crefrange{eqn:transmittance_exponential}{eqn:freeflight_exponential}, and the attenuation coefficient satisfies \cref{eqn:reciprocity_sigma}, as required for reciprocity.
We discuss reversibility and prove \cref{the:coefficient} in \cref{sec:proof_theorem}. 

\vspace{-0.50em}
\subsection{Anisotropy}\label{sec:anisotropy}
\vspace{-0.50em}

Returning to \cref{eqn:coeff_markov}, we can rewrite it as the product:%
%
%
\vspace{-18pt}
\begin{equation}\label{eqn:coeff_markov_reorg}
	\mymathboxnt{intBlueLL}{\coeff_\delta\paren{\point,\direction}} = \mymathbox{intTealL}{intTealLL}{$\equiv \density\paren{\point}$}{\frac{\norm{\nabla \vacancy\paren{\point}}}{\vacancy\paren{\point}}} \cdot \mymathbox{intGrayL}{intGrayLL}{$\equiv \projecteddelta\paren{\point, \direction}$}{\vphantom{\frac{\norm{\nabla \vacancy\paren{\point}}}{\vacancy\paren{\point}}}\abs{\direction \cdot \normal\paren{\point}}},
	\vspace{-5pt}
\end{equation}
where $\normal\paren{\point} \equiv \nicefrac{\nabla\vacancy\paren{\point}}{\norm{\nabla\vacancy\paren{\point}}}$ is the \emph{unit normal} of the level set of $\vacancy$ passing through $\point$. We compare \cref{eqn:coeff_markov_reorg} to the attenuation coefficient expressions for \emph{anisotropic} stochastic microparticle geometry by \citet[Equation (11)]{jakob2010anisotropy} and \citet[Equation (2)]{heitz2015sggx}. As in those works, we can decompose the attenuation coefficient as the product of an isotropic \emph{density} $\density\paren{\point}$ and an anisotropic \emph{projected area} $\projecteddelta\paren{\point, \direction}$.%
\footnote{The projected area and density should include multiplicative factors $\mathrm{A}_{\vacancy}\paren{\point}$ and  $\nicefrac{1}{\mathrm{A}_{\vacancy}\paren{\point}}$, respectively, where $\mathrm{A}_{\vacancy}\paren{\point}\equiv \norm{\nabla \vacancy\paren{\point}}\ud x \ud y \ud z$ is the differential area of the tangent plane of the level set of $\vacancy$ at $\point$. These factors cancel out in \cref{eqn:coeff_markov_reorg}, so we omit them to simplify notation, at the cost of our expressions \emph{appearing} to have incorrect units.}
%
Intuitively:
\begin{enumerate*}
	\item The density $\density\paren{\point}$ increases as vacancy $\vacancy\paren{\point}$ decreases; thus the ray termination probability through $\point$ increases the more likely $\point$ is to be occupied. 
	\item The projected area $\projecteddelta\paren{\point}$ models foreshortening as a ray of direction $\direction$ encounters a surface patch of normal $\normal\paren{\point}$; at \emph{grazing angles} ($\abs{\direction\cdot\normal\paren{\point}}=0$) the patch is invisible, whereas at \emph{normal incidence} ($\abs{\direction\cdot\normal\paren{\point}} = 1$) it is maximally visible, corresponding to zero and maximal, respectively, ray termination probability. 
\end{enumerate*}

This anisotropic behavior mimics deterministic surfaces, and thus is suitable for points $\point$ likely to lie on the surface of a stochastic opaque solid (i.e., $\vacancy\paren{\point} \approx \nicefrac{1}{2}$). However, points $\point$ that are likely inside the solid (i.e., $\vacancy\paren{\point} \approx 0$, respectively) should behave \emph{isotropically}: rays passing through them at different directions should terminate with the same probability. To model these different behaviors, inspired from microflake models for microparticle geometry~\citep{jakob2010anisotropy,heitz2015sggx}, we generalize our definitions of $\projecteddelta$ and $\coeff_\delta$.%
\footnote{We follow \citet{jakob2010anisotropy} and use normalized distributions of normals. 
}
\begin{dfn}[label={def:anisotropy}]{Attenuation coefficient for opaque solids}{anisotropy}
	We associate with each point $\point \in \R^3$ a \emph{distribution of normals} $\ndf_{\point} : \Sph^2 \to \R_{\ge 0}$ that satisfies $\int_{\Sph^2} \ndf_{\point}\paren{\normalalt} \ud \normalalt = 1$. Then, we define at $\point$ the \emph{projected area} for any direction $\direction \in \Sph^2$ as the expected foreshortening:
	\vspace{-5pt}
	\begin{equation}
		\mymathboxnt{intGrayLL}{\projected_{\ndf}\paren{\point, \direction} \equiv \int_{\Sph^2} \abs{\direction \cdot \normalalt} \ndf_{\point}\paren{\normalalt} \ud \normalalt}, \label{eqn:projected}
		\vspace{-5pt}
	\end{equation}
	the \emph{density} as
	\vspace{-5pt}
	\begin{equation}
		\mymathboxnt{intTealLL}{\density\paren{\point} \equiv \frac{\norm{\nabla \vacancy\paren{\point}}}{\vacancy\paren{\point}}}, \label{eqn:density}
		\vspace{-5pt}
	\end{equation}
	and the \emph{generalized attenuation coefficient} as the product:
	\vspace{-15pt}
	\begin{equation}\label{eqn:coeff_generalized}
		\mymathboxnt{intBlueLL}{\coeff\paren{\point,\direction}\vphantom{\projected_{\ndf}}} \equiv \mymathboxnt{intTealLL}{\density\paren{\point}\vphantom{\projected_{\ndf}}} \cdot \mymathboxnt{intGrayLL}{\projected_{\ndf}\paren{\point, \direction}}.
	\end{equation}
\end{dfn}

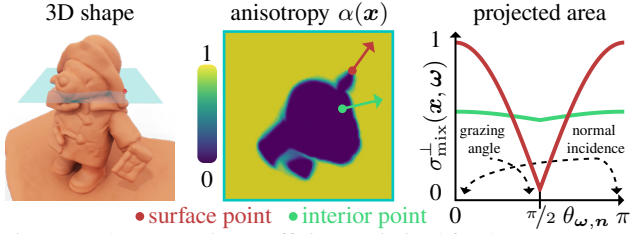
\begin{figure}[t]
	\centering
	\includegraphics[width=\columnwidth]{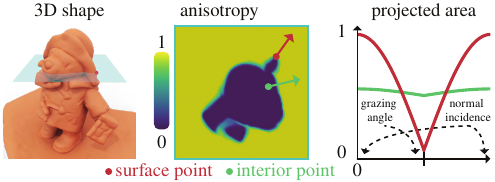}
	\put(-110pt,79.5pt){{\small $\anisotropy\paren{\point}$}}
	\put(-25pt,2.5pt){{\small $\theta_{\direction,\normal}$}}
	\put(-5pt,2.5pt){{\small $\pi$}}
	\put(-39pt,2.5pt){{\small $\nicefrac{\pi}{2}$}}
	\put(-79pt,22pt){{\small \rotatebox{90}{$\projected_{\mathrm{mix}}\paren{\point, \direction}$}}}
	\vspace{-1em}
	\caption{The attenuation coefficient optimized for the \textsc{bear} scene in BlendedMVS behaves as anticipated by our theory: isotropically in the object interior, and anisotropically near its surface.}
	\label{fig:anisotropy}
	\vspace{-2em}
\end{figure}

\vspace{-0.25em}
For $\ndf_{\point,\delta}\paren{\normalalt} \equiv \delta\paren{\normalalt - \normal\paren{\point}}$, the projected area reduces to $\abs{\direction \cdot \normalalt}$, explaining the notation $\coeff_{\delta}, \projecteddelta$ in \crefrange{eqn:coeff_markov}{eqn:coeff_markov_reorg}. By contrast, for the uniform distribution $\ndf_{\point,\mathrm{unif}}\paren{\normalalt} \equiv \nicefrac{1}{4\pi}$, the projected area becomes $\projected_{\mathrm{unif}}\paren{\point, \direction} \equiv \nicefrac{1}{2}$; then both the projected area and attenuation coefficient are \emph{isotropic}. Definition~\ref{def:anisotropy} allows behaviors between these two extremes, e.g., by using a linear mixture distribution of normals $\ndf_{\point,\mathrm{mix}}\paren{\normalalt} \equiv \anisotropy\paren{\point} \ndf_{\point,\delta}\paren{\normalalt} + \paren{1 - \anisotropy\paren{\point}} \ndf_{\point,\mathrm{unif}}\paren{\normalalt}$ and corresponding projected area:
\vspace{-12pt}
\begin{align}
	&\mymathboxnt{intGrayLL}{\projected_{\mathrm{mix}}\paren{\point, \direction} \equiv \anisotropy\paren{\point} \projecteddelta\paren{\point,\direction} + \paren{1-\anisotropy\paren{\point}} \projected_{\mathrm{unif}}\paren{\point, \direction}} \nonumber \\
	&\qquad\qquad\quad \mymathboxnt{intGrayLL}{= \anisotropy\paren{\point} \abs{\direction \cdot \normal\paren{\point}} + \frac{1-\anisotropy\paren{\point}}{2}}.\label{eqn:projected_mixture}
	\vspace{-20pt}
\end{align}
The \emph{anisotropy parameter} $\anisotropy\paren{\point}\in\bracket{0,1}$ continuously interpolates between fully anisotropic ($\anisotropy\paren{\point} = 1$) and fully isotropic ($\anisotropy\paren{\point} = 0$) projected area. Making this parameter spatially varying allows adapting the anisotropy behavior at different parts of an opaque solid, e.g., more anisotropic near its boundary, more isotropic in its interior (\cref{fig:anisotropy}). We discuss additional choices for the distribution of normals $\ndf$ and associated projected area $\projected_{\ndf}$ in \cref{sec:illumination,sec:quantitative}.

\vspace{-0.5em}
\subsection{Stochastic implicit surfaces}\label{sec:implicit}
\vspace{-0.5em}

Definitions~\ref{def:opaque_solid} and~\ref{def:stochastic_solid} define a (stochastic) solid through the binary indicator function, because the indicator (respectively vacancy) function at a point $\point$ is the \emph{minimal} information we need to determine visibility (respectively transmittance) for a ray that passes through $\point$. However, it is common practice to define a solid through a non-binary scalar field, which provides richer information about the solid and its surface~\citep{osher2005level}. We explore this case next. 

\begin{definition}\label{def:implicit}
	We define an \emph{implicit function} $\implicit : \R^3 \to \R$ as a real scalar field, and associate with it an indicator function $\indicator\paren{\point} \equiv \ind_{\curly{\implicit\paren{x} \le 0}}$ and corresponding solid $\solid$. If $\implicit$ is also a \emph{random} field, then we define the pointwise \emph{cumulative distribution function} $\cdf_{\implicit\paren{\point}}$, \emph{probability density function} $\pdf_{\implicit\paren{\point}}$, and \emph{mean implicit function} $\meanimplicit\paren{\point}$ as, respectively:
	\vspace{-10pt}
	\begin{align}
		\cdf_{\implicit\paren{\point}} \paren{\imval} &\equiv \Prob\curly{\implicit\paren{\point} \le \imval},\: \imval\in\R, \label{eqn:cdf}\\
		\pdf_{\implicit\paren{\point}}\paren{\imval} &\equiv \frac{\ud \,\cdf_{\implicit\paren{\point}}\paren{\imval}}{\ud \imval},\: \imval \in \R, \label{eqn:pdf} \\
		\meanimplicit\paren{\point} &\equiv \E{\implicit\paren{\point}} = \int_{-\infty}^{+\infty} \imval\cdot \pdf_{\implicit\paren{\point}}\paren{\imval} \ud \imval. \label{eqn:mean}
	\end{align}
	\vspace{-22pt}
\end{definition}

From Definition~\ref{def:implicit}, the stochastic solid $\solid$ is an \emph{excursion set} of the random field $\implicit$ \citep[Chapter 1]{adler2010geometry}, with its surface at the zero-level set of $\implicit$, its interior where $\implicit < 0$, and its exterior elsewhere. Such excursion sets have been extensively studied in applied mathematics, especially when $\implicit$ is a \emph{Gaussian process}, i.e., its (joint) distribution at one or more points is Gaussian \citep[Appendix]{adler2010geometry}. \citet{sellan2022stochastic,sellan2023neural} recently proposed using excursion sets of Gaussian processes as a point-based stochastic implicit surface representation (\cref{sec:pointcloud}). Extending our theory to excursion sets of various stochastic implicit functions $\implicit$ will allow us to provide stochastic geometry interpretations for previous volumetric representations for opaque solids \citep{yariv2021volume,wang2021neus}.

To this end, we specialize to stochastic implicit functions with a \emph{symmetry} property: At each $\point$, $\implicit\paren{\point}$ equals, up to a spatially varying shift $\meanimplicit\paren{\point}$ and spatially constant scale $s > 0$, a zero-mean, unit-variance, and \emph{symmetric} random variable---that is, with a PDF $\pdfunc : \R \to \R_{\ge 0}$ and CDF $\cdfunc : \R \to \bracket{0, 1}$ that satisfy, for all $\imval \in \R$ \citep{casella2021statistical},
\begin{equation}\label{eqn:symmetric}
	\pdfunc\paren{\imval} = \pdfunc\paren{-\imval} \text{ and }\cdfunc\paren{\imval} = 1 - \cdfunc\paren{-\imval}. 
\end{equation}
Such a CDF $\cdfunc$ is a \emph{sigmoid} function \citep{han1995influence,glorot2011deep} whose exact shape depends on the probability distribution. Common symmetric distributions include the Gaussian, logistic, and Laplace (in their zero-mean, unit-variance versions), giving rise to the error, logistic, and Laplace (respectively) sigmoid functions. The symmetry property implies for $\implicit$:
\begin{align}
	\pdf_{\implicit\paren{\point}}\paren{\imval} &= \pdfunc\paren{s\paren{\imval - \meanimplicit\paren{\point}}},\label{eqn:specialize0}\\
	\cdf_{\implicit\paren{\point}}\paren{\imval} &= \cdfunc\paren{s\paren{\imval - \meanimplicit\paren{\point}}}.\label{eqn:specialize}
\end{align}
%
We prove in \cref{sec:proof_proposition} the following proposition.
\begin{prp}[label={pro:implicit}]{Stochastic implicit geometry}{implicit}
	We assume a stochastic implicit function $\implicit\paren{\point}$ satisfying \crefrange{eqn:specialize0}{eqn:specialize}. Then, the occupancy and vacancy for the associated stochastic solid $\solid$ equal:
	\begin{align}
		\occupancy\paren{\point} &= \Prob\curly{\implicit\paren{\point} \le 0} = \cdfunc\paren{-s\meanimplicit\paren{\point}},\label{eqn:occupancy_implicit}\\
		\vacancy\paren{\point} &= \Prob\curly{\implicit\paren{\point} > 0} = \cdfunc\paren{s\meanimplicit\paren{\point}}. \label{eqn:vacancy_implicit}
	\end{align}
	If the stochastic solid $\solid$ also satisfies the conditions of \cref{thm:coefficient}, then the attenuation coefficient equals:
	\vspace{-10pt}
	\begin{equation}\label{eqn:coeff_implicit}
		\!\mymathboxnt{intBlueLL}{\coeff\paren{\point,\direction}} \!=\! \mymathbox{intTealL}
		{intTealLL}
		{$\equiv\density\paren{\point}$}
		{\frac{s\pdfunc\paren{s\meanimplicit\paren{\point}}\norm{\nabla \meanimplicit\paren{\point}}}{\cdfunc\paren{s\meanimplicit\paren{\point}}}} \!\cdot\!
		\mymathboxnt{intGrayLL}
		{\projected_{\ndf}\paren{\point,\direction}},
		\vspace{-5pt}
	\end{equation}
	with $\projected_{\ndf}$ as in \cref{eqn:projected}. 
\end{prp}

\vspace{-0.25em}
\Cref{pro:implicit} completes our volumetric representation, which we summarize in \cref{fig:theory}. Notably, the modeling choice to use an implicit function impacts the density $\density$ (through the vacancy $\vacancy$), but \emph{not} the projected area $\projected_{\ndf}$. 
 To help intuition, we discuss the behavior of different quantities.

\paragraph{Vacancy.} The vacancy $\vacancy$ (\cref{eqn:vacancy_implicit}) is a sigmoidal transform of the expected value $\meanimplicit$ of the stochastic implicit function $\implicit$ (\cref{eqn:mean}). This agrees with intuition: a large positive value of $\meanimplicit\paren{\point}$ (i.e., high probability that $\point$ is \emph{outside} the solid) results in $\vacancy\paren{\point} \approx 1$; a large negative value of $\meanimplicit\paren{\point}$ (i.e., high probability that $\point$ is \emph{inside} the solid) results in $\vacancy\paren{\point} \approx 0$; and $\meanimplicit\paren{\point} = 0$ (i.e., equal probability that $\point$ is inside or outside the solid) results in $\vacancy\paren{\point} = \nicefrac{1}{2}$. 

\paragraph{Attenuation coefficient and free-flight distribution.} The behaviors of the attenuation coefficient $\coeff$ (\cref{eqn:coeff_implicit}) and free-flight distribution $\freeflight$ (\cref{eqn:freeflight_exponential}) are easier to understand if we consider points on a ray along which the mean implicit $\meanimplicit$ monotonically decreases with a constant gradient. Then, the attenuation coefficient monotonically increases along the ray (i.e., as we move from points likely in the exterior to points likely in the interior of the solid). The free-flight distribution is maximal where along the ray $\meanimplicit\paren{\point} = 0$ (i.e., at the point equally likely to be inside or outside the solid), and decreases as the magnitude of $\meanimplicit$ increases (i.e., at points highly likely to be inside or outside the solid).

\paragraph{Scale and uncertainty.} The scale $s$ controls the width of the sigmoid $\cdfunc$, and thus how fast the vacancy $\vacancy$ transitions from 0 to 1 as the mean implicit function $\meanimplicit$ increases---the larger $s$ is, the narrower $\cdfunc$ becomes, and the faster $\vacancy$ changes. We can interpret this behavior by noticing that, from \crefrange{eqn:specialize0}{eqn:specialize}, $s$ is the inverse of the pointwise standard deviation of $\implicit$. As $s$ increases, the standard deviation decreases, and thus: 
\begin{enumerate*}
	\item the pointwise PDF $\pdf_{\implicit}$ becomes more concentrated around its mean $\meanimplicit$ (i.e., the stochastic implicit function $\implicit$ becomes more certain); 
	\item the vacancy $\vacancy$ and occupancy $\occupancy$ become closer to the binary functions $\ind_{\curly{\meanimplicit > 0}}$ and  $\ind_{\curly{\meanimplicit \le 0}}$ (i.e., the random indicator function $\indicator$ becomes more certain). 
	\item the free-flight distribution $\freeflight$ becomes closer to a Dirac delta function centered at the zero-level set of $\meanimplicit$ (i.e., free-flight distances become more certain).
\end{enumerate*}
%


\vspace{-1em}
\section{Relationship to prior work}\label{sec:prior}
\vspace{-0.5em}

\begin{table}[t]
	\centering
	\caption{Classification of previous and new volumetric representations for opaque solids using our theory (Figure~\ref{fig:theory}).\label{tab:prior}}
	\vspace{-0.75em}
	\setlength{\aboverulesep}{0pt}
	\setlength{\belowrulesep}{0pt}
	\small
	\begin{tabularx}{\linewidth}
	{@{} l >{\columncolor{intTealLL}}l >{\columncolor{intGrayLL}}X}
	\toprule
	\rowcolor{white}
	\textbf{method}	& \makecell[l]{\textbf{\textcolor{intTealL}{implicit function}} \\ \textbf{\textcolor{intTealL}{distribution $\mathbf{\Psi}$}}}	& \makecell[l]{\textbf{\textcolor{intGrayL}{distribution}} \\ \textbf{\textcolor{intGrayL}{of normals $\mathbf{D}$}}} \\
	\midrule
	VolSDF	& Laplace	& uniform \\
	NeuS	& logistic	& delta (with ReLU) \\
	\makecell[l]{NeuS with \\ cosine annealing}	& logistic & \makecell[l]{mixture (with ReLU, \\ constant anisotropy)} \\
	\textbf{ours}	& Gaussian & \makecell[l]{mixture (with \\
	$\point$-varying anisotropy)} \\
	\bottomrule
	\end{tabularx}
	\vspace{-3em}
\end{table}

Our theory provides a volumetric representation for opaque solids that permits various probabilistic modeling choices, e.g., selecting a distribution of normals (\cref{eqn:projected}) and an implicit function distribution (\cref{eqn:coeff_implicit}). We use our theory to explain and compare volumetric representations in prior work as versions of our representation corresponding to specific choices for these distributions (\cref{tab:prior}), albeit with critical caveats that our theory addresses.

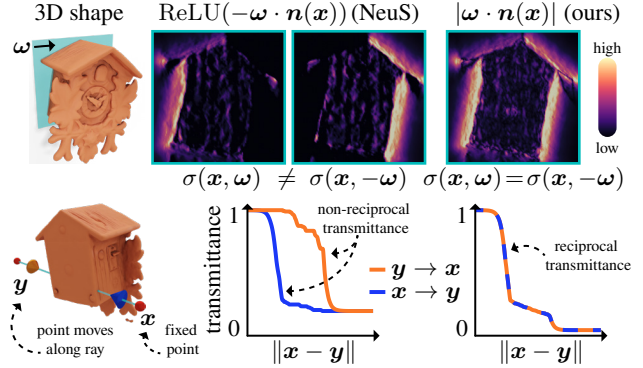
\begin{figure}[t]
	\centering
	\includegraphics[width=\columnwidth]{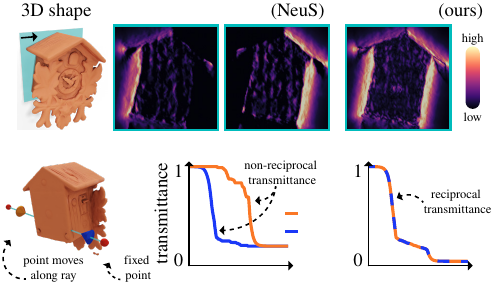}
	\put(-80pt,66pt){{\small $\coeff\paren{\point,\direction} \!=\! \coeff\paren{\point,-\direction}$}}
	\put(-123pt,66pt){{\small $\coeff\paren{\point,-\direction}$}}
	\put(-171pt,66pt){{\small $\coeff\paren{\point,\direction}$}}
	\put(-135pt,66pt){{\small $\neq$}}
	\put(-68pt,128.5pt){{\small $\abs{\direction\cdot\normal\paren{\point}}$}}
	\put(-180pt,128.5pt){{\small $\mathrm{ReLU}\paren{-\direction\cdot\normal\paren{\point}}$}}
	\put(-235pt,116pt){{\small $\direction$}}
	\put(-187pt,13pt){{\small $\point$}}
	\put(-235pt,26pt){{\small $\pointalt$}}
	\put(-92pt,32pt){{\small $\pointalt\to\point$}}
	\put(-92pt,23pt){{\small $\point\to\pointalt$}}
	\put(-137.5pt,0.5pt){{\small $\norm{\point-\pointalt}$}}
	\put(-51.5pt,0.5pt){{\small $\norm{\point-\pointalt}$}}
	\vspace{-0.75em}
	\caption{When optimizing for the \textsc{clock} scene in BlendedMVS using NeuS, the $\mathrm{ReLU}$ term leads to attenuation coefficient (top) and transmittance (bottom) values that violate reciprocity. By contrast, using our representation leads to reciprocal results.}
	\label{fig:reciprocity}
	\vspace{-1.5em}
\end{figure}

\paragraph{NeuS.} Most closely related to our work is the NeuS volumetric representation by \citet[Equation (10)]{wang2021neus}. Using our notation, their extinction coefficient equals:
\vspace{-5pt}
\begin{align}
	&\!\!\!\mymathboxnt{intBlueLL}{\coeff_{\mathrm{NeuS}}\paren{\point, \direction}} \equiv \nonumber \\
	&\!\!\!\mymathboxnt{intTealLL}{\frac{s \pdfunc_{\mathrm{logistic}}\paren{s \meanimplicit\paren{\point}} \norm{\nabla\meanimplicit\paren{\point}}}{\cdfunc_{\mathrm{logistic}}\paren{s \meanimplicit\paren{\point}}}} \!\cdot\! \mymathboxnt{intGrayLL}{\mathrm{ReLU}\paren{-\direction\cdot\normal\paren{\point}}},\label{eqn:coeff_neus}
	\vspace{-15pt}
\end{align}
where: 
\begin{enumerate*}
	\item $\pdfunc_{\mathrm{logistic}}$ and $\cdfunc_{\mathrm{logistic}}$ are the PDF and CDF, respectively, for the zero-mean, unit-variance \emph{logistic distribution}; and 
	\item the mean implicit function $\meanimplicit$ is parameterized as a neural field that, during training, is also regularized to approximate a signed distance function (i.e., satisfy $\norm{\nabla \meanimplicit\paren{\point}} \approx 1$).
\end{enumerate*}
Comparing \cref{eqn:coeff_neus} with our \cref{eqn:coeff_implicit,eqn:coeff_markov_reorg}, we see that the NeuS model is close to our model with Dirac delta distribution of normals $\ndf_{\point,\delta}$, specialized to specific choices for the pointwise distribution $\pdfunc$ and mean implicit function $\meanimplicit$, with an important difference: it replaces the Dirac delta projected area $\projecteddelta\paren{\point,\direction}\equiv\abs{\direction\cdot\normal\paren{\point}}$ with the anisotropic term $\mathrm{ReLU}\paren{-\direction\cdot\normal\paren{\point}} \equiv \max\paren{0, -\direction \cdot \normal\paren{\point}}$, effectively clipping $\coeff_{\mathrm{NeuS}}$ to zero when a ray is traveling outwards (towards larger vacancy or mean implicit values). Unfortunately, this choice has the consequence that the attenuation coefficient $\coeff_{\mathrm{NeuS}}$ violates the reciprocity requirement of \cref{eqn:reciprocity_sigma}, resulting in a physically implausible volumetric representation. We visualize this issue in \cref{fig:reciprocity}. As we discuss in \cref{sec:experiments}, violation of reciprocity additionally negatively impacts 3D reconstruction quality.

Lastly, we prove in \cref{sec:proof_logistic} that using the logistic distribution for $\implicit\paren{\point}$ simplifies \cref{eqn:coeff_implicit}:
\vspace{-5pt}
\begin{align}
	&\mymathboxnt{intBlueLL}{\coeff_{\mathrm{logistic}}\paren{\point, \direction}} = \nonumber \\
	&\mymathboxnt{intTealLL}{s\Psi_{\mathrm{logistic}}\paren{-s\meanimplicit\paren{\point}}\norm{\nabla\meanimplicit\paren{\point}}\vphantom{\projected_{\ndf}\paren{\point,\direction}}} \cdot \mymathboxnt{intGrayLL}{\projected_{\ndf}\paren{\point,\direction}},\label{eqn:coeff_implicit_logistic}
	\vspace{-20pt}
\end{align}
and analogously for $\coeff_{\mathrm{NeuS}}\paren{\point, \direction}$ in \cref{eqn:coeff_neus}. We use this observation as we discuss VolSDF next.

\paragraph{VolSDF.} Another closely related model is the VolSDF model by~\citet[Equations (2)--(3)]{yariv2021volume}. Using our notation:
%
\vspace{-3pt}
\begin{equation}
	\!\!\!\!\mymathboxnt{intBlueLL}{\coeff_{\mathrm{VolSDF}}\paren{\point, \direction}} \!\equiv\! \mymathboxnt{intTealLL}{s\Psi_{\mathrm{Laplace}} \paren{-s\meanimplicit\paren{\point}}\norm{\nabla\meanimplicit\paren{\point}}},\label{eqn:coeff_volsdf}
	\vspace{-2pt}
\end{equation}
where $\cdfunc_{\mathrm{Laplace}}$ is the CDF for the zero-mean, unit-variance \emph{Laplace distribution}; and the mean implicit function $\meanimplicit\paren{\point}$ is parameterized as in NeuS. Comparing to \cref{eqn:coeff_implicit_logistic}, we make two observations about the VolSDF model: 
\begin{enumerate*}
	\item It uses the uniform distribution of normals $\ndf_{\point,\mathrm{unif}}$ and isotropic (constant) projected area $\projected_{\mathrm{unif}}$, thus the attenuation coefficient becomes \emph{isotropic}. 
	\item It uses a density $\density$ equal to that in \cref{eqn:coeff_implicit_logistic} after replacing the logistic with the Laplace CDF. However, this replacement is \emph{not} equivalent to modeling $\implicit\paren{\point}$ as a Laplace random variable. This is because the simplified expression of \cref{eqn:coeff_implicit_logistic} is correct for only the logistic distribution, whereas the Laplace distribution requires the full expression of \cref{eqn:coeff_implicit}. Consequently, the VolSDF model uses an incorrect density term.
\end{enumerate*}


\paragraph{Cosine annealing.} The official NeuS implementation~\citep[\texttt{models/renderer.py\#L232-L235}]{NeuSCodebase} uses \emph{cosine annealing}---transitioning from isotropy to anisotropy as optimization progresses---to improve convergence. This means replacing the anisotropic term in \cref{eqn:coeff_neus} with:
\vspace{-3pt}
\begin{equation}
	\mymathboxnt{intGrayLL}{\anisotropy \cdot \mathrm{ReLU}\paren{-\direction\cdot\normal\paren{\point}} + \frac{1-\anisotropy}{2}},\label{eqn:coeff_annealing}
	\vspace{-3pt}
\end{equation}
and changing the \emph{global} anisotropy parameter $\anisotropy$ from 0 towards 1 using a predetermined annealing schedule. Comparing to \cref{eqn:projected_mixture}, our theory explains this heuristic as using a mixture distribution of normals, with the caveat that NeuS also replaces $\projecteddelta$ with the reciprocity-violating $\mathrm{ReLU}$ term, as we noted earlier. Additionally, NeuS uses a spatially constant anisotropy parameter $\anisotropy$ that cannot capture the qualitatively different foreshortening behavior of surface versus interior and exterior points (\cref{sec:experiments}).

\paragraph{Scale optimization and adaptive shells.} NeuS and VolSDF optimize the scale $s$, which typically increases as optimization progresses. Our theory explains this behavior as decreasing uncertainty of the stochastic geometry and convergence towards deterministic geometry (binary vacancy). 

The \emph{adaptive shells} representation by \citet{zian2023adaptive} modifies the NeuS representation to use a spatially varying scale $s\paren{\point}$. Our theory explains this choice as spatially varying pointwise standard deviation, and thus uncertainty, for the stochastic implicit function $\implicit\paren{\point}$. In \cref{sec:proof_proposition} we explain how to modify the density $\density$ in \cref{eqn:coeff_implicit} to account for spatially varying scale $s\paren{\point}$.

\paragraph{Other approaches.} In the appendix we discuss other approaches for modeling stochastic solid geometry, such as occupancy networks \citep{mescheder2019occupancy,niemeyer2020differentiable} and discretization approaches \citep{bhotika2002probabilistic,tulsiani2017multi,oechsle2021unisurf} in \cref{sec:otherprior}, as well as stochastic implicit surfaces using point clouds \citep{sellan2022stochastic} in \cref{sec:pointcloud}.
\begin{table}[t]
	\centering
	\caption{Chamfer distance statistics on the DTU and NeRF Realistic Synthetic datasets. We provide the full tables in \cref{sec:quantitative}. \label{tab:quantitative}}
	\vspace{-0.75em}
	\begin{tabularx}{\linewidth}{@{} X >{\centering\arraybackslash}X >{\centering\arraybackslash}X >{\centering\arraybackslash}X @{}}
		\toprule
		DTU & VolSDF & NeuS & \textbf{ours} \\
		\midrule
		mean & 1.84 & 2.17 & \textbf{1.57} \\
		median & 1.74 & 1.99 & \textbf{1.56} \\
		\midrule
		NeRF RS & VolSDF & NeuS & \textbf{ours} \\
		\midrule
		mean & 0.252 & 0.201 & \textbf{0.113} \\
		median & 0.100 & 0.085 & \textbf{0.057} \\
		\bottomrule
	\end{tabularx}
	\vspace{-1em}
\end{table}

\begin{table}
	\centering
	\caption{We use chamfer distance statistics on the DTU dataset for an ablation study. We provide the full tables in \cref{sec:quantitative}. \label{tab:ablation}}
	\vspace{-0.75em}
	\stackengine{0pt}{%
		\begin{tabularx}{\linewidth}{@{} X >{\centering\arraybackslash}X >{\centering\arraybackslash}X >{\centering\arraybackslash}X @{}}
			\toprule
			$\cdfunc$ model & logistic & Laplace & \textbf{Gaussian} \\
			\midrule
			mean & 1.98 & 1.96 & \textbf{1.78} \\
			median & 1.86 & 1.92 & \textbf{1.74} \\
			\midrule
		\end{tabularx}
	}{%
		\begin{tabularx}{\linewidth}{@{} X >{\centering\arraybackslash}X >{\centering\arraybackslash}X >{\centering\arraybackslash}X >{\centering\arraybackslash}X @{}}
			$\ndf$ model & delta ($\mathrm{ReLU}$) & delta & mixture (const.) & \textbf{mixture (var.)} \\
			\midrule
			mean & 2.17 & 1.98 & 1.97 & \textbf{1.75} \\
			median & 1.99 & 1.86 & 1.85 & \textbf{1.59} \\
			\bottomrule
		\end{tabularx}
	}{U}{c}{F}{F}{S}
	\vspace{-1.8em}
\end{table}

\vspace{-1em}
\section{Experimental evaluation}\label{sec:experiments}
\vspace{-0.5em}

Our theory provides a framework for systematically \emph{correcting} volumetric representations for opaque solids in prior work (using reciprocal projected area and correct density), and \emph{designing} new representations (using different distributions for the implicit function and normals). These changes are straightforward to implement within existing neural rendering pipelines. We do so within a simplified version of NeuS \citep{wang2021neus} and perform extensive experiments on multi-view 3D reconstruction tasks. The goal of these experiments is not state-of-the-art performance, but an exploration of the design space and an evaluation of volumetric representations within an equal framework. Our results show that, despite their simple nature, the changes our theory suggests greatly improve performance, qualitatively and quantitatively. Overall, our experiments demonstrate the utility of a rigorous theory of volumetric representations for opaque solids.

In this section, we summarize our experiments and findings. We discuss implementation details and list complete numerical results in \cref{sec:quantitative}. Lastly, we provide interactive visualizations and code on the project website.

\begin{figure}[t]
	\centering
	\includegraphics[width=\linewidth]{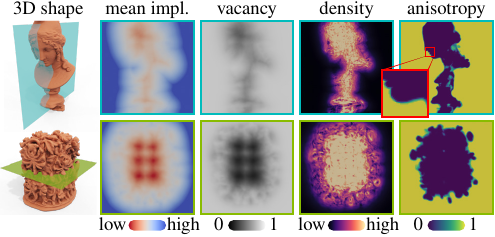}
	\vspace{-2.25em}
	\caption{Visualization of shape and key quantities of our volumetric representation learned for scenes in the BlendedMVS dataset.}
	\label{fig:fields}
	\vspace{-2em}
\end{figure}

\begin{figure*}[t]
	\centering
	\includegraphics[width=\textwidth]{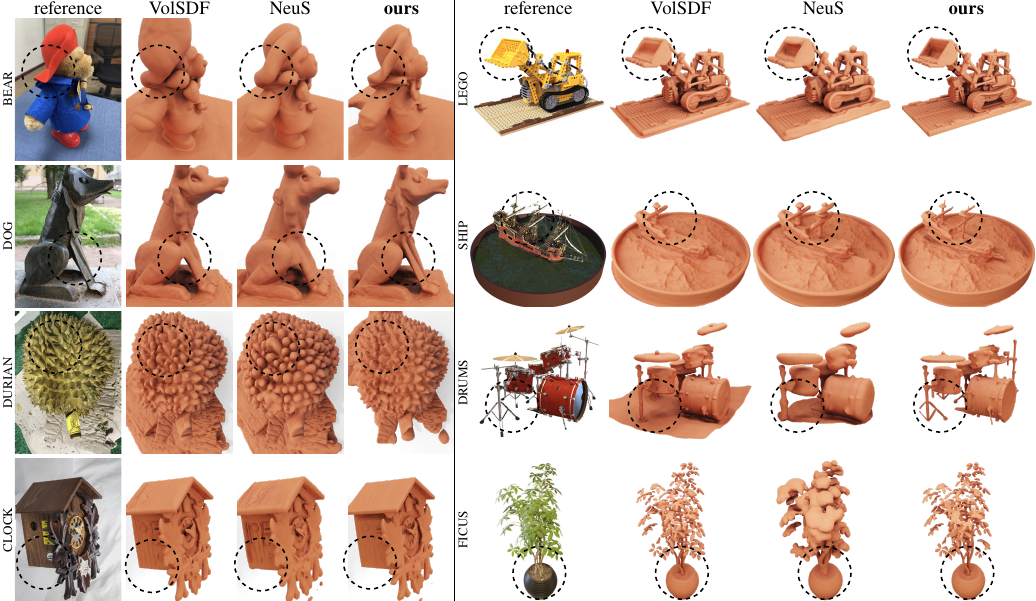}
	\vspace{-2em}
	\caption{Qualitative comparisons on the BlendedMVS (left) and NeRF Realistic Synthetic (right) datasets. The dashed circles indicate areas of interest. We provide interactive visualizations of results on the complete datasets on the project website.}
	\label{fig:qualitative}
	\vspace{-1.75em}
\end{figure*}

\paragraph{Comparison to prior work.} We evaluate our best performing volumetric representation against those of NeuS \citep{wang2021neus} and VolSDF \citep{yariv2021volume}. \Cref{tab:prior} summarizes the three representations. We use three datasets for evaluation: DTU \citep{DTU}, BlendedMVS \citep{BlendedMVS}, and NeRF realistic synthetic (NeRF RS) \citep{mildenhall2021nerf}. 

\Cref{tab:quantitative} and \cref{fig:qualitative} show quantitative and qualitative results. We observe the following:
\begin{enumerate*}
	\item Our representation performs the best across all datasets, both quantitatively and qualitatively. Qualitatively, the improvements are more pronounced in BlendedMVS and NeRF RS---whose scenes have more complex geometry, materials, and background---than in DTU---whose simpler scenes are reconstructed well by all representations.
	\item Our representation learns meaningful scalar fields (\cref{fig:fields}) for the mean implicit function $\meanimplicit$, vacancy $\vacancy$, density $\density$, and anisotropy $\anisotropy$. The mean implicit function and vacancy lend themselves to downstream surface processing tasks (e.g., mesh extraction \citep{lorensenMarchingCubes}). 
	\item The use of a spatially varying anisotropy $\anisotropy$ allows our representation to model the qualitatively different behaviors of points $\point$ on the surface ($\anisotropy\paren{\point} \approx 1$, i.e., strongly anisotropic) versus in the interior ($\anisotropy\paren{\point} \approx 0$, i.e., isotropic). By contrast, VolSDF and NeuS require \emph{all} points $\point$---surface or interior---to have either isotropic ($\anisotropy\paren{\point} \coloneqq 0$) or perfectly anisotropic ($\anisotropy\paren{\point} \coloneqq 1$), respectively, behavior.
\end{enumerate*}

\paragraph{Design and ablation.} We use the DTU dataset to evaluate design choices for the implicit function distribution $\cdfunc$ and distribution of normals $\ndf$ in \cref{eqn:coeff_implicit,eqn:projected}, respectively. This evaluation also serves as an ablation study for our best performing representation in \cref{tab:prior}. \Cref{tab:ablation} shows the results. We observe the following:
\begin{enumerate*}
	\item For $\cdfunc$, using the Gaussian distribution (i.e., a Gaussian process \citep{sellan2022stochastic}) outperforms using the Laplace (as in VolSDF) or logistic (as in NeuS) distributions.
	\item For $\ndf$, using the spatially varying mixture distribution (\cref{eqn:projected_mixture}, to adapt to surface versus interior points) outperforms using the spatially constant mixture distribution (as in cosine annealing) or the delta distribution. 
	\item When using the delta distribution, including a reciprocity-violating $\mathrm{ReLU}$ term (\cref{eqn:coeff_neus}, as in NeuS) underperforms not doing so. This result highlights that enforcing reciprocity not only ensures physical plausibility, but also improves reconstruction performance.
\end{enumerate*}

\vspace{-1.1em}
\section{Conclusion}\label{sec:conclusion}
\vspace{-0.5em}

We have taken first steps towards formalizing volumetric representations for opaque solids, using stochastic geometry theory. Our results should be of theoretical and practical interest: On the theory side, they help explain why volumetric neural rendering can reconstruct solid geometry, and justify previous volumetric representations for it. On the practice side, they provide a toolbox for the design of physically-plausible volumetric representations that greatly improve performance.
We hope that our results will motivate further research along both theory and practice thrusts. 

For the theory thrust, our theory is far from a complete formalism of volumetric representations for solid geometry. We highlight three important shortcomings that deserve further investigation. First, revisiting \cref{eqn:volume}, we focused on the \emph{geometry} of opaque solids, but neglected their \emph{global illumination} effects. In \cref{sec:illumination}, we briefly examine how geometry and global illumination must be coupled 
to ensure reciprocity. However, this topic requires further investigation.
Second, we excluded \emph{(semi-)transparent} solids, where interior points may be visible to each other (violating Definition~\ref{def:opaque_solid}). 
Developing volumetric representations for such solids will allow modeling complex reflective-refractive appearance. Third, we focused on \emph{exponential} transport because it has served as a convenient approximation for most prior work. However, both empirical evidence \citep{vicini2021non} and stochastic geometry theory suggest that exponentiality may not be a suitable assumption for opaque solids. Indeed, the excursion sets of Gaussian processes typically have free-flight distributions (also known as the \emph{first-passage times} of the Gaussian processes) that are \emph{non-exponential} \citep{aalen2001understanding}. As \citet[Section 5.1]{sellan2023neural} point out, characterizing these distributions requires reasoning about the \emph{spatial covariance} structure of the Gaussian process \citep[Appendix]{adler2010geometry}.

For the practice thrust, 
we have done only limited evaluation of how different volumetric representations interplay with different algorithms for free-flight estimation and sampling (\cref{sec:implementation}). The theoretical investigation and empirical evaluation of such algorithms will be critical for optimizing volumetric neural rendering pipelines for solid geometry.

\paragraph{Acknowledgments.} We thank Aswin Sankaranarayanan for helpful discussions. This work was supported by NSF awards 1900849, 2008123, NSF Graduate Research Fellowship DGE2140739, an NVIDIA Graduate Fellowship for Miller, and a Sloan Research Fellowship for Gkioulekas.

{\small
\bibliographystyle{ieeenat_fullname}
\bibliography{probsurf}
}

\clearpage
\setcounter{page}{1}
\maketitlesupplementary

\appendix

\section{Volumetric representation of point clouds}\label{sec:pointcloud}

In this section, we discuss how our theory relates to prior work on modeling point clouds, and how their combination enables volume rendering of point clouds.

\paragraph{Stochastic Poisson surface reconstruction (SPSR).} \citet{sellan2022stochastic} propose a stochastic model for opaque solids as excursion sets of Gaussian processes conditioned on point clouds. Using our notation, their model corresponds to \crefrange{eqn:occupancy_implicit}{eqn:vacancy_implicit} as
\begin{equation}
	\vacancy_{\mathrm{SPSR}}\paren{\point} = 1 - \occupancy_{\mathrm{SPSR}}\paren{\point} \equiv \cdfunc_{\mathrm{Gaussian}}\paren{s\paren{\point}\meanimplicit\paren{\point}}, \label{eqn:vacancy_spsr}
\end{equation}
where $\cdfunc_{\mathrm{Gaussian}}$ is the CDF for the zero-mean, unit-variance Gaussian distribution; and the mean implicit function $\meanimplicit\paren{\point}$ and scale $s\paren{\point}$ are parameterized as functionals of a point cloud. For volume rendering, they use the occupancy $\occupancy_{\mathrm{SPSR}}$ in place of the attenuation coefficient $\coeff$ \citep[Section 8.2]{sellan2022stochastic}. Our theory disambiguates the occupancy and corresponding attenuation coefficient, which from \cref{eqn:coeff_generalized} (assuming perfectly anisotropic projected area) becomes:
\vspace{-5pt}
\begin{equation}
	\mymathboxnt{intBlueLL}{\coeff_{\mathrm{SPSR}}\paren{\point,\direction}\vphantom{\projected_{\ndf}}} \equiv \mymathboxnt{intTealLL}{\frac{\norm{\nabla \vacancy_{\mathrm{SPSR}}\paren{\point}}}{\vacancy_{\mathrm{SPSR}}\paren{\point}}} \cdot \mymathboxnt{intGrayLL}{\abs{\direction \cdot \normal\paren{\point}}}.\label{eqn:coeff_pc}
	\vspace{-5pt}
\end{equation}

\begin{figure*}[t]
	\centering
	\includegraphics[width=\linewidth]{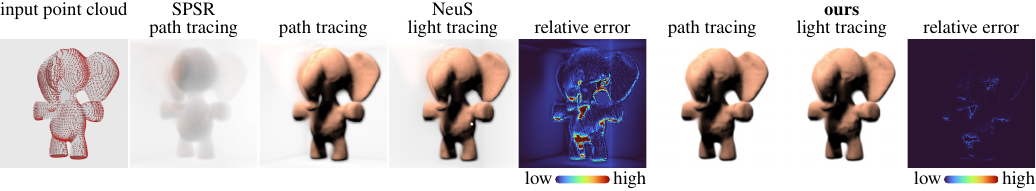}
	\vspace{-2.3em}
	\caption{Comparison of point-cloud volume renderings on a scene from SPSR~\citep{sellan2022stochastic}. Using the vacancy $\vacancy_{\mathrm{SPSR}}$ in place of the attenuation coefficient as SPSR suggests results in images that do not convey the appearance of an opaque solid. Using a non-reciprocal attenuation coefficient $\coeff_{\mathrm{SPSR,NeuS}}$ as NeuS suggests results in significant differences between images from different rendering algorithms. Using the reciprocal attenuation coefficient $\coeff_{\mathrm{SPSR}}$ our theory suggests results in accurate images under both rendering algorithms.}
	\label{fig:point_cloud}
	\vspace{-1.8em}
\end{figure*}

\paragraph{Volume rendering of point clouds.} To highlight the importance of disambiguating vacancy and the attenuation coefficient, we use single-bounce volume rendering to synthesize images of a point cloud scene from SPSR \citep{sellan2022stochastic} under point light illumination. In \cref{fig:point_cloud}, we compare the images resulting from using the attenuation coefficient our theory predicts in \cref{eqn:coeff_pc}, versus using the vacancy in \cref{eqn:vacancy_spsr} in place of the attenuation coefficient. Using the attenuation coefficient results in images that more accurately convey the appearance of the opaque solid underling the point cloud.

We use the same scene for an additional experiment that highlights the importance of enforcing reciprocity in volumetric representations of opaque solids. We consider variant of \cref{eqn:coeff_pc} that replaces the perfectly anisotropic projected area with the $\mathrm{ReLU}$-based one from NeuS \citep{wang2021neus}:
\vspace{-5pt}
\begin{align}
	&\mymathboxnt{intBlueLL}{\coeff_{\mathrm{SPSR,NeuS}}\paren{\point,\direction}\vphantom{\projected_{\ndf}}} \equiv \nonumber \\
	&\mymathboxnt{intTealLL}{\frac{\norm{\nabla \vacancy_{\mathrm{SPSR}}\paren{\point}}}{\vacancy_{\mathrm{SPSR}}\paren{\point}}} \cdot \mymathboxnt{intGrayLL}{\mathrm{ReLU}\paren{-\direction \cdot \normal\paren{\point}}}.\label{eqn:coeff_pc_neus}
	\vspace{-5pt}
\end{align}
We use the attenuation coefficients of \cref{eqn:coeff_pc,eqn:coeff_pc_neus} to synthesize images with two volume rendering algorithms: volumetric path tracing (trace rays starting from the camera) and volumetric light tracing (trace rays starting from the light source) \citep{veach1998robust}. Under a physically-plausible, reciprocal volumetric representation, both algorithms should produce identical images. We show in \cref{fig:point_cloud} that using our reciprocal attenuation coefficient in \cref{eqn:coeff_pc} produces images that are a lot closer to each other (up to discretization and noise artifacts) than those using the non-reciprocal attenuation coefficient in \cref{eqn:coeff_pc_neus} that NeuS suggests.
\section{Global illumination representation}\label{sec:illumination}

We briefly discuss the global illumination term in the volume rendering equation \labelcref{eqn:volume}. In \emph{deterministic} light transport algorithms, global illumination appears as the outgoing radiance in the conservation of radiance equation. From the \emph{reflection equation}, the outgoing radiance equals:
\begin{align}\label{eqn:reflection}
	&L_\mathrm{o}\paren{\point, \direction_\mathrm{o}} = \nonumber \\
	&\int_{\Hemi^2\paren{\point}}\!\!\!\! \brdf\paren{\point, \normal\paren{\point}, \direction_\mathrm{o}, \direction_\mathrm{i}} L_\mathrm{i}\paren{\point, \direction_\mathrm{i}} \inprod{\direction_\mathrm{i}}{\normal\paren{\point}} \ud \direction_\mathrm{i},
\end{align}
where  $\Hemi^2\paren{\point} \equiv \curly{\direction \in \Sph^2 : \direction\cdot\normal_{\point} \ge 0}$ is the positive hemisphere and $\brdf$ is the bidirectional distribution reflection function (BRDF) at point $\point$. In \emph{volumetric} light transport algorithms, global illumination appears as the conditional expectation of the outgoing radiance in the volume rendering equation \labelcref{eqn:volume}. Dropping the distinction between expected and actual radiance to ease notation, from the \emph{in-scattering equation} \citep[Section 3.1]{heitz2015sggx}, the outgoing radiance equals:
\begin{equation}\label{eqn:inscattering}
	L_\mathrm{o}\paren{\point, \direction_\mathrm{o}} = \albedo\paren{\point}\int_{\Sph^2} \pf\paren{\point, \direction_\mathrm{o}, \direction_\mathrm{i}} L_\mathrm{i}\paren{\point, \direction_\mathrm{i}} \ud \direction_\mathrm{i},
\end{equation}
where $\albedo \in \bracket{0,1}$ is the albedo and $\pf$ is the phase function at point $\point$. \citet[Section 3]{jakob2010anisotropy} prove that reciprocal \emph{exponential} transport requires that the attenuation coefficient $\coeff$ and phase function $\pf$ satisfy for all directions $\direction_\mathrm{o}, \direction_\mathrm{i}$:
\begin{equation}\label{eqn:reciprocity_pf}
	\coeff\paren{\point,\direction_\mathrm{o}} \pf\paren{\point,\direction_\mathrm{o},\direction_\mathrm{i}} = \coeff\paren{\point,\direction_\mathrm{i}} \pf\paren{\point,\direction_\mathrm{i},\direction_\mathrm{o}}.
\end{equation}
Lastly, \citet[Section 5]{heitz2015sggx} show that, when the attenuation coefficient has the form of \cref{eqn:coeff_generalized} in Definition~\ref{def:anisotropy}, the phase function $\pf$ relates to the BRDF $\brdf$ of the underlying deterministic geometry as:
\begin{align}\label{eqn:microflake}
	&\pf\paren{\point,\direction_\mathrm{o},\direction_\mathrm{i}} = \frac{1}{\projected_{\ndf}\paren{\point,\direction_\mathrm{o}}} \nonumber \\
	&\qquad\cdot \int_{\Sph^2} \brdf\paren{\point, \normalalt, \direction_\mathrm{o}, \direction_\mathrm{i}} \abs{\direction_{\mathrm{o}} \cdot \normalalt} \ndf_{\point}\paren{\normalalt} \ud \normalalt.
\end{align}
\Cref{eqn:microflake} expresses the phase function as the expected value of the \emph{foreshortened} BRDF with respect to the distribution of normals $\ndf$ used also to define the projected area $\projected_{\ndf}$ in \cref{eqn:projected}. We make three observations:
\begin{enumerate*}
	\item The phase function in \cref{eqn:microflake} satisfies the reciprocity relation in \cref{eqn:reciprocity_pf}. 
	\item \citet{heitz2015sggx} derived \cref{eqn:microflake} in the context of microflake models for stochastic microparticle geometry. The similarity between the attenuation coefficient in our theory (Definition~\ref{def:anisotropy}) and that in microflake models allows us to apply \cref{eqn:microflake} also for stochastic solid geometry.
	\item As in the microflake model for stochastic microparticle geometry, we can use different BRDFs $\brdf$ in \cref{eqn:microflake} (e.g., specular or Lambertian) to derive corresponding phase functions for the stochastic solid geometry. Then, different distribution of normals $\ndf$ correspond to different levels of ``roughness'' of the underlying geometry, in analogy with the distribution of normals in microfacet BRDF models (e.g., GGX \citep{walter2007microfacet} or Oren-Nayar \citep{oren1993diffuse} models for rough specular or Lambertian, respectively, reflectance). However, reciprocity requires using the same distribution of normals for both the phase function and attenuation coefficient, and thus jointly constrains geometry and global illumination in the volume rendering equation \labelcref{eqn:volume}.
\end{enumerate*}

\paragraph{Integrated directional encoding.} Neural rendering pipelines typically represent the global illumination in \cref{eqn:volume} as a neural field with positional and directional inputs. This representation makes it difficult to exactly enforce \cref{eqn:inscattering,eqn:microflake}. \citet{verbin2022ref} showed how to approximately do so with an \emph{integrated directional encoding} that computes moments of directional inputs with respect to the distribution of normals describing surface roughness. \Cref{eqn:microflake} suggests that we can combine our volumetric representation with integrated directional encoding, by using the same distribution of normals $\ndf$ for both.%
\footnote{A caveat to this approach is that, whereas our Definition~\ref{def:anisotropy} and \cref{eqn:microflake} use $\ndf$ as a distribution of \emph{normals}, the integrated directional encoding of \citet{verbin2022ref} uses it as a distribution of \emph{reflected directions}. This discrepancy is analogous to the difference between the Blinn-Phong and Phong BRDFs, which randomize normals (or equivalently, half-way directions) versus reflected directions, respectively \citep{blinn1977models}. \citet[Appendix]{ramamoorthi2001signal} show that, for the von-Mises Fisher distribution that \cref{eqn:projected_vmf} and the integrated directional encoding use, this discrepancy introduces only a small approximation error.}
This approach links the geometry and global illumination terms in the volume rendering equation \labelcref{eqn:volume}, as reciprocity requires.

In particular, \citet{verbin2022ref} use a distribution of normals $\ndf_{\point}$ that equals the von-Mises Fisher distribution with center at the normal $\normal\paren{\point}$ and concentration parameter $\kappa\paren{\point} \equiv \nicefrac{\anisotropy\paren{\point}}{1-\anisotropy\paren{\point}}$, where $\anisotropy\paren{\point} \in \bracket{0,1}$ is a spatially varying anisotropy parameter.%
\footnote{\citet{verbin2022ref} parameterize the von-Mises Fisher distribution using an \emph{unbounded} roughness parameter $\rho \in \leftinc{0,\infty}$, which we replace with our \emph{bounded} anisotropy parameter $\anisotropy \equiv \nicefrac{1}{\rho\paren{\point} + 1}$.}
Using \cref{eqn:projected}, the corresponding projected area equals:
\vspace{-5pt}
\begin{equation}
	\mymathboxnt{intGrayLL}{\projected_{\mathrm{vMF}}\paren{\point, \direction} \approx \frac{\anisotropy\paren{\point} + 1}{2}\abs{\direction\cdot\normal\paren{\point}}^{\anisotropy\paren{\point}}}. \label{eqn:projected_vmf}
	\vspace{-5pt}
\end{equation}
The proof of \cref{eqn:projected_vmf} is exactly analogous to that by \citet[Section 7.2]{han2007frequency}, and uses the same approximations as in the derivation of the equations for the integrated directional encoding \citep[Equation (8), Appendix A]{verbin2022ref}. It is worth comparing $\projected_{\mathrm{vMF}}$ in \cref{eqn:projected_vmf} to $\projected_{\mathrm{mix}}$ and $\projected_{\mathrm{SGGX}}$ in \cref{eqn:projected_mixture,eqn:projected_sggx}: all three projected area terms become fully anisotropic and isotropic at the limit values $\anisotropy = 1$ and $\anisotropy = 0$, respectively, and produce different interpolations between these extremes at intermediate values.

\begin{figure}[t]
	\includegraphics[width=\columnwidth]{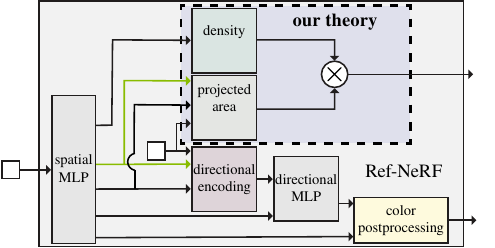}
	\put(-141pt,93.5pt){{\scriptsize Eq. \labelcref{eqn:density}}}
	\put(-141pt,57.3pt){{\scriptsize Eq. \labelcref{eqn:projected_vmf}}}
	\put(-66.75pt,86.5pt){{\scriptsize Eq. \labelcref{eqn:coeff_implicit}}}
	\put(-105pt,92pt){{\scriptsize $\density$}}
	\put(-105pt,59pt){{\scriptsize $\projected_{\mathrm{vMF}}$}}
	\put(-7pt,81.5pt){{\scriptsize $\coeff$}}
	\put(-7pt,11.5pt){{\scriptsize $c$}}
	\put(-234.5pt,36pt){{\scriptsize $\point$}}
	\put(-165pt,44.5pt){{\scriptsize $\direction$}}
	\put(-190pt,61pt){{\scriptsize $\meanimplicit$}}
	\put(-190pt,42.5pt){{\scriptsize \textcolor{greenarrow}{$\rho$}}}
	\put(-190pt,30pt){{\scriptsize $\hat{\normal}$}}
	\put(-190pt,17pt){{\scriptsize $\theta$}}
	\put(-190pt,7pt){{\scriptsize $c_d$, $s$}}
	\vspace{-0.75em}
	\caption{Visualization of our modifications to Ref-NeRF \citep{verbin2022ref}.}
	\vspace{-1em}
	\label{fig:refnerf}
\end{figure}

\begin{figure}[t]
	\centering
	\includegraphics[width=\columnwidth]{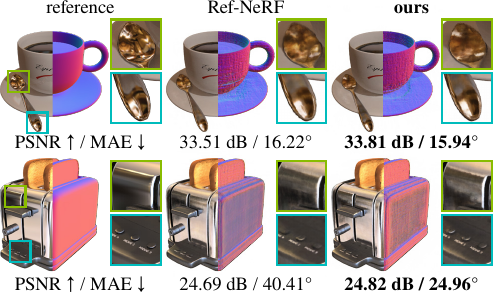}
	\vspace{-2em}
	\caption{Qualitative results for the Shiny Blender dataset.}\label{fig:nvs}
	\vspace{-1em}
\end{figure}

\paragraph{Novel-view synthesis experiments.} We demonstrate the importance of linking representations for geometry and global illumination in the volume rendering equation \cref{eqn:volume} through \emph{proof-of-concept} experiments on novel-view synthesis. We use Ref-NeRF \citep{verbin2022ref} as a baseline, and modify its architecture to use our attenuation coefficient in \cref{eqn:coeff_implicit} with projected area as in \cref{eqn:projected_vmf} (\cref{fig:refnerf}). Importantly, our modification uses the same roughness $\rho$ (equivalently, anisotropy $\anisotropy$) for both the projected area and integrated directional encoding. The qualitative and quantitative results in \cref{fig:nvs} and \cref{tab:nvs_psnr,tab:nvs_mae} show that our modifications improve normal and appearance predictions.

\paragraph{Implementation details.} We briefly discuss important implementation details for our multi-view synthesis experiments. We build on the Ref-NeRF implementation in the Multi-NeRF codebase~\cite{multinerf2022}. We modify the normal loss to compare with the level-set normals of the vacancy, $\nicefrac{\nabla \vacancy}{\norm{\nabla \vacancy}}$, instead of the attenuation coefficient, $\nicefrac{\nabla \coeff}{\norm{\nabla \coeff}}$---our theory shows that the two are \emph{not} equal (\cref{eqn:coeff_generalized}) and the former should better represent the underlying geometry.

\begin{table}
	\centering
	\caption{PSNR values on the Shiny Blender dataset.}\label{tab:nvs_psnr}
	\vspace{-1em}
	\begin{tabularx}{\linewidth}{@{} X >{\centering\arraybackslash}X >{\centering\arraybackslash}X @{}}
		\toprule
		Shiny Blender & Ref-NeRF & \textbf{ours} \\
		\midrule
		\textsc{teapot}  & \textbf{44.07}  & 43.35\\
		\textsc{toaster} & 24.69           & \textbf{24.82}\\
		\textsc{car}     & \textbf{29.07}  & 28.84\\
		\textsc{ball}    & 32.48           & \textbf{32.79}\\
		\textsc{coffee}  & 33.51           & \textbf{33.81}\\
		\textsc{helmet}  & 29.31           & \textbf{30.31}\\
		\midrule
		mean & 32.19 & \textbf{32.32} \\
		median & 30.89 & \textbf{31.55} \\
		\bottomrule
	\end{tabularx}
	\vspace{-1em}
\end{table}

\begin{table}
	\centering
	\caption{Mean average error (MAE) values for learned normals on the Shiny Blender dataset.}\label{tab:nvs_mae}
	\vspace{-1em}
	\begin{tabularx}{\linewidth}{@{} X >{\centering\arraybackslash}X >{\centering\arraybackslash}X @{}}
		\toprule
		Shiny Blender & Ref-NeRF & \textbf{ours} \\
		\midrule
		\textsc{teapot}  & 48.48         & \textbf{42.86}\\
		\textsc{toaster} & 40.41         & \textbf{24.96}\\
		\textsc{car}     & 21.68         & \textbf{18.06}\\
		\textsc{ball}    & 98.94         & \textbf{38.18}\\
		\textsc{coffee}  & 16.22         & \textbf{15.94}\\
		\textsc{helmet}  & 50.62         & \textbf{21.46}\\
		\midrule
		mean & 46.06 & \textbf{26.91}\\
		median & 44.45 & \textbf{23.21} \\
		\bottomrule
	\end{tabularx}
	\vspace{-1em}
\end{table}

Multi-NeRF takes advantage of the large memory capacity of TPUs and uses large batch sizes during training. We had access to only GPUs with \qty{16}{\giga\byte} of memory, so we had to reduce the batch size from the default $2^{14}$ to $2^9$ rays per batch. We also followed a suggestion in the Github repository of Multi-NeRF to reduce the learning rate when using smaller batch sizes. Instead of scheduling the learning rate to decay from $2\times 10^{-3}$ to $2\times 10^{-5}$, we use a schedule from $5\times 10^{-4}$ to $5\times 10^{-6}$. These settings lead to worse performance than what Ref-NeRF reports \citep{verbin2022ref}. Therefore, for a fair comparison, we run Ref-NeRF with our training settings. We use the default \qty{250}{\kilo\nothing} iterations, which on an NVIDIA V100 Tensor Core GPU takes 18 hours.


\section{Relationship to other prior work}\label{sec:otherprior}

\paragraph{Occupancy networks.} \citet{mescheder2019occupancy} and \citet{niemeyer2020differentiable} propose an alternative to volume rendering for stochastic opaque solid models. They define an occupancy and vacancy model as in \cref{eqn:occupancy_implicit} similar to NeuS,
\begin{equation}
	\vacancy_{\mathrm{ONet}}\paren{\point} \equiv \cdfunc_{\mathrm{logistic}}\paren{s\meanimplicit\paren{\point}}. \label{eqn:occupancy_onet}
\end{equation}
where the mean implicit function $\meanimplicit\paren{\point}$ is a (trainable) neural field. However, instead of converting this occupancy to an attenuation coefficient $\coeff$ for volume rendering, they define what we term the \emph{maximum a posteriori} (MAP) opaque solid $\solid_{\mathrm{MAP}} \equiv \curly{\point \in \R^3 : \vacancy\paren{x} \le \nicefrac{1}{2}}$. This approach is equivalent to setting at each point $\point$ the random indicator function $\indicator\paren{\point}$ equal to its MAP estimate $\hat{\indicator}_{\mathrm{MAP}}\paren{\point} \equiv \ind_{\curly{\vacancy\paren{x} \le \nicefrac{1}{2}}}$---hence the term MAP opaque solid.

\citet{mescheder2019occupancy} proceed to render $\solid_{\mathrm{MAP}}$ using \emph{deterministic} intersection queries and rendering algorithms. This approach is analogous to setting, during deterministic rendering, the free-flight distance $\casting_{\point, \direction}$ equal to its MAP estimate $\casting_{\point, \direction,\mathrm{MAP}} \equiv \argmax\displaystyle_{\distance \in \leftinc{0,+\infty}} \freeflight_{\point, \direction}\paren{\distance}$. By contrast, volume rendering (\cref{eqn:volume}) computes the expectation over the free-flight distance  $\casting_{\point, \direction}$. By using point estimation instead of expectation, the MAP approach discards uncertainty information available in $\occupancy\paren{\point}$ about the scene. Additionally, recent work has demonstrated the benefits of volumetric representations for novel-view synthesis~\citep{mildenhall2021nerf,verbin2022ref,barron2021mip} and 3D reconstruction~\citep{oechsle2021unisurf,wang2021neus,yariv2021volume}, motivating our focus on volumetric representations. 

\paragraph{Discretization approaches.} Another approach for deriving volumetric representations for opaque solids is to discretize a ray and define the free-flight distribution using discrete occlusion events~\citep{bhotika2002probabilistic,tulsiani2017multi,oechsle2021unisurf}. We specialize this discussion to exponential transport. Along each ray, these approaches are discrete-time approximations to the continuous-time binary Markov process $\indicator_{\point, \direction}\paren{\distance}$ underlying our model~\citep{norris1998markov}. Alternatively, but equivalently, they correspond to a piecewise-constant approximation to the attenuation coefficient $\sigma$ associated with the indicator function, as we explain below. 

We consider a segment $\bracket{\distance_0,\distance_N}$ along a ray $\ray_{\point,\direction}\paren{\distance}$, which we break into $N$ segments $\bracket{\distance_n, \distance_{n+1}},\; n=0,\dots,N-1$, such that $\distance_0 < \distance_1 < \dots < \distance_N$. (These segments can have unequal lengths.) The probability that a ray will \emph{not} terminate within segment $\bracket{\distance_n, \distance_{n+1}}$ equals:
\begin{align}\label{eqn:discrete_vacancy}
	\Vacancy_n &\equiv \transmittance_{\point, \direction}\paren{\distance_n,\distance_{n+1}} \\ 
	&= \exp\curly{-\int_{\distance_n}^{\distance_{n+1}} \coeff\paren{\ray_{\point,\direction}\paren{\distanceVariable},\direction} \ud \distanceVariable},
\end{align}
where we used Definition~\ref{def:general_defs} and expanded the notation for transmittance in \cref{eqn:transmittance_ray,eqn:transmittance_exponential}.
The quantity $\Vacancy_n$ is a discrete analogue of the vacancy $\vacancy$ of the continuous-time Markov process $\indicator_{\point, \direction}\paren{\distance}$: it equals the probability that $\indicator_{\point,\direction}\paren{\distance} = 0$ for all $\distance \in \bracket{\distance_n, \distance_{n+1}}$, and thus all points along the ray segment are vacant. We also define a discrete occupancy as $\Occupancy_n \equiv 1 - \Vacancy_n$. With this notation, assuming Markovianity, the probability that a ray terminates at segment $n$ after traveling through the previous segments---i.e., the discrete analogue of the free-flight distribution---equals:
\begin{equation}\label{eqn:discrete_free_flight}
	\Freeflight_n \equiv \Occupancy_n \prod_{m = 0}^{n-1} \Vacancy_m = \paren{1 - \Vacancy_n} \prod_{m = 0}^{n-1} \Vacancy_m.
\end{equation}
Lastly, if we approximate the attenuation coefficient on the segment $\bracket{\distance_n, \distance_{n+1}}$ as a constant $\bar{\coeff}_n$, then from Equation~\eqref{eqn:discrete_vacancy} the discrete vacancy becomes:
\begin{equation}\label{eqn:exponential_discrete_vacancy}
	\Vacancy_n = \exp\curly{-\bar{\coeff}_n\cdot\abs{\distance_{n+1}-\distance{n}}}.
\end{equation}
\Crefrange{eqn:discrete_free_flight}{eqn:exponential_discrete_vacancy} correspond exactly to the model for ray tracing through discretized opaque geometry in previous literature~\citep{bhotika2002probabilistic,tulsiani2017multi,ulusoy2015towards,oechsle2021unisurf}. As we discuss in \cref{sec:implementation}, they are also equivalent to the common procedure for numerically \emph{approximating} the free-flight distribution when volume rendering continuous opaque geometry \citep{max1995optical,mildenhall2021nerf}. 

Importantly, most previous discretization-based approaches model not the (piecewise-linear) attenuation coefficient $\bar{\coeff}$ in \cref{eqn:exponential_discrete_vacancy}, but directly the discrete vacancy $\Vacancy_n$ (or occupancy $\Occupancy_n$). They do so using either a grid~\citep{bhotika2002probabilistic,tulsiani2017multi,ulusoy2015towards}, or a neural field~\citep{oechsle2021unisurf}---i.e., $\Vacancy_n \coloneqq \mathrm{neural\_net}\paren{\ray_{\point, \direction}\paren{\distance_n}}$. 

Unfortunately, directly modeling the discrete vacancy in this way does not account for the ray-segment length term $\abs{\distance_{n+1}-\distance{n}}$ in \cref{eqn:exponential_discrete_vacancy}, which is problematic when considering segments of varying length (e.g., in multi-resolution approaches, or when ray discretization is non-uniform). As an example, we consider the discrete free-flight distribution $\Freeflight_n$ in \cref{eqn:discrete_free_flight} for the same ray $\ray_{\point, \direction}\paren{\distance}$, first at discretization resolution $\Delta \distance$, then at a finer resolution $\nicefrac{\Delta \distance}{2}$. In the latter case, $\Freeflight_n$ will equal the product of twice as many $\Vacancy_n$ terms as in the first case. If these terms are directly provided by, e.g., a neural field that does not account for the change in ray-segment length as resolution changes, then increasing discretization resolution results in a much lower value for the discrete free-flight distribution of the same ray. By contrast, we can elide this problem by using the neural field to output the attenuation coefficient $\bar{\coeff}_n$, then computing the vacancy terms $\Vacancy_n$ through \cref{eqn:discrete_vacancy} while accounting for varying ray-segment lengths.
\section{Implementation details}\label{sec:implementation}

We discuss implementation details for our 3D reconstruction experiments. Our volumetric neural rendering pipeline builds on top of the NeuS pipeline \citep{wang2021neus}.

\paragraph{Quadrature algorithm.} An important algorithmic component of volumetric rendering pipelines is the numerical integration (i.e., quadrature) along a ray of the free-flight distribution and expected radiance terms in the volume rendering equation \labelcref{eqn:volume}. Quadrature algorithms work in two stages: first, they sample locations along the ray; second, they approximate the free-flight distribution and expected radiance along the ray segments between samples. Several quadrature algorithms have appeared in both graphics~\citep{georgiev2019integral,kettunen2021unbiased,novak2014residual,nimier2022unbiased,kutz2017spectral} and vision~\citep{mildenhall2021nerf,verbin2022ref,barron2021mip,yariv2021volume,oechsle2021unisurf,wang2021neus}. In particular, the volumetric representations in prior work we discussed in \cref{sec:prior} all use their own quadrature  algorithms. Unfortunately, this makes it difficult to experimentally compare different volumetric representations on equal terms: empirically, we have found that performance differences between different representations are often primarily due to the different underlying quadrature algorithms, rather than the different attenuation coefficient expressions. Therefore, to ensure a fair and informative comparison, in our experiments we use for all volumetric representations the same quadrature algorithm, which combines a sampling technique inspired by \citet{oechsle2021unisurf} with the numerical approximation by \citet{max1995optical} (\crefrange{eqn:discrete_free_flight}{eqn:exponential_discrete_vacancy}). We have found that this quadrature algorithm performs well for all representations we tested.

\begin{figure}[t]
	\centering
	\includegraphics[width=\columnwidth]{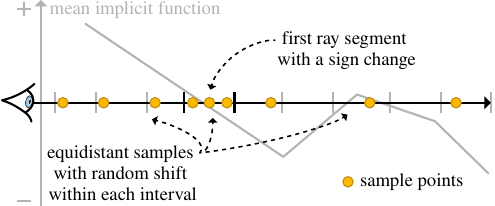}
	\put(-129pt,91.5pt){{\small \textcolor{graysampling}{$\meanimplicit$}}}
	\vspace{-0.5em}
	\caption{Visualization of our importance sampling algorithm, whoch concentrates points within the first ray segment at whose endpoints the mean implicit function $\meanimplicit$ has opposite signs.}
	\label{fig:sampling}
	\vspace{-1.5em}
\end{figure}

The sampling component of our quadrature algorithm, which we visualize in \cref{fig:sampling}, works as follows: We first intersect a ray with a bounding sphere of the scene. Then, we divide the ray into 1024 segments, evaluate the mean implicit function $\meanimplicit$ at the endpoints of each segment, and check for sign changes between the endpoints. The free-flight distribution $\freeflight$ is concentrated within the segment containing the first sign change of $\meanimplicit$, and \citet{oechsle2021unisurf} suggest placing most samples in this segment. We follow their suggestion and sample a total of 64 points in three sets---one-third of the sample points within the segment with the first sign change, another third in the part of the ray before this segment, and the remaining third in the part of the ray after. If there is no segment with a sign change, we allocate the entire set of 64 samples along the entire ray. For each of the three sample sets, we select the sample locations within their corresponding ray intervals using the equidistant sampling comb approach of \citet{kettunen2021unbiased}: we deterministically place the samples at equidistant intervals, then shift them all by the same uniformly-sampled random offset. 

\paragraph{Network architecture.} Our pipeline mimics NeuS \citep{wang2021neus}, using two multi-layer perceptrons (MLPs) to encode the mean implicit function and volume emission, with the addition of another small MLP to encode the anisotropy parameter.

The mean implicit function MLP has 8 hidden layers with 256 neurons per layer and $\mathrm{SoftPlus}$ activations with $\beta=100$. A skip connection combines the output of the fourth layer with the original input. It takes as input the spatial position, and outputs the mean implicit function value and a 256-dimensional feature vector.

The volume emission MLP has 4 hidden layers with 256 neurons per layer and $\mathrm{ReLU}$ activations. It takes as input the spatial position, viewing direction, unit normal of the isosurface of the vacancy function, and the 256-dimensional feature vector from the mean implicit function MLP.

The anisotropy MLP has 1 hidden layer with 256 neurons and sigmoid activation. It takes as input the 256-dimensional feature vector from the mean implicit function MLP.

All positional and directional inputs to the MLPs use frequency encoding \citep{mildenhall2021nerf}, with 6 frequencies for positional and 4 frequencies for directional inputs. As in NeuS, all linear layers use weight normalization~\cite{salimans2016weightNormalization} to stabilize training.

\paragraph{Background radiance field.} To enable surface reconstruction on unbounded scenes, such as those in the BlendedMVS dataset, we follow the approach of NeuS \citep{wang2021neus} and VolSDF \citep{yariv2021volume}: We perform surface reconstruction only inside a bounding sphere, and treat its exterior as a neural radiance field. We use the \emph{inverted sphere parameterization} from NeRF++ \citep{kaizhang2020nerfpp} for the exterior of the sphere. To learn the background radiance field, we sample along each ray an additional 32 points on the exterior of the bounding sphere, uniformly with respect to inverse distance in the range $\leftinc{0, \nicefrac{1}{R}}$, where $R=3$ is the bounding sphere radius. As in VolSDF, we do not learn a background radiance field for the DTU scenes.

\paragraph{Regularization and initialization.} We follow NeuS \citep{wang2021neus} and use eikonal regularization \citep{gropp2020eikonalReg}, and geometric initialization \citep{atzmon2020geometricInit} for the mean implicit function $\meanimplicit$.

\paragraph{Training details.} Our training protocol is the same as in NeuS \citep{wang2021neus}. We train using the ADAM optimizer \citep{kingma2015adam} for \qty{300}{\kilo\nothing} iterations with 512 rays per batch. During the first \qty{5}{\kilo\nothing} iterations, we control the learning rate with a linear warmup from $0$ to $5\times 10^{-4}$; for the remaining iterations, we use a cosine-decay schedule until the learning rate reaches a minimum of $2.5\times 10^{-5}$. Training on an NVIDIA V100 Tensor Core GPU training takes 11 hours without the background radiance field, and 12 hours with it.
\section{Additional quantitative results}\label{sec:quantitative}

We provide complete versions of \crefrange{tab:quantitative}{tab:ablation} in \crefrange{tab:dtu_full}{tab:nerf_full} (for \cref{tab:quantitative}) and \crefrange{tab:density_full}{tab:projected_full} (for \cref{tab:ablation}).

In \cref{tab:projected_full}, we evaluate an additional option for the distribution of normals: We use the \emph{SGGX distribution} that \citeauthor{heitz2015sggx} propose for stochastic microparticle geometry---specifically its \emph{surface-like} version \citep[Equations (13)--(14)]{heitz2015sggx}. Using our notation and a reparameterization in terms of a (spatially varying) anisotropy parameter $\anisotropy\paren{\point} \in \bracket{0,1}$, this distribution and corresponding projected area equal:
\begin{align}
	&\!\!\!\ndf_{\point,\mathrm{SGGX}}\paren{\normalalt} \!\equiv\! \frac{1}{\ndf\paren{\anisotropy\paren{\point}}\paren{1\! -\! \anisotropy\paren{\point}^2\abs{\normalalt \cdot \normal\paren{\point}}^2}^2}, \label{eqn:ndf_sggx} \\
	&\!\!\!\mymathboxnt{intGrayLL}{\projected_{\mathrm{SGGX}}\paren{\point, \direction} \equiv} \nonumber \\
	&\!\!\!\mymathboxnt{intGrayLL}{\frac{1}{\coeffnorm\paren{\anisotropy\paren{\point}}} \sqrt{\anisotropy\paren{\point}^2 \abs{\direction \cdot \normal\paren{\point}}^2 + (1 - \anisotropy\paren{\point}^2)}}, \label{eqn:projected_sggx}
	\vspace{-10pt}
\end{align}
with normalization coefficients
\begin{align}
	\ndf\paren{\anisotropy} &\equiv \frac{\anisotropy + \arctanh\paren{\anisotropy} - \anisotropy^2\arctanh\paren{\anisotropy}}{2 \anisotropy \pi},\\
	\coeffnorm\paren{\anisotropy} &\equiv 1 + \paren{\frac{1}{\anisotropy} -  \anisotropy} \arcsinh\paren{\frac{\anisotropy}{\sqrt{1-\anisotropy^2}}}.
\end{align}
Comparing the SGGX projected area of \cref{eqn:projected_sggx} to the mixture one in \cref{eqn:projected_mixture}, we see that---the normalization coefficients aside---they behave similarly: $\projected_{\mathrm{SGGX}}$ is a sum of squared terms, whereas $\projected_{\mathrm{mix}}$ is a sum of linear terms. At the limits $\anisotropy = 1$ and $\anisotropy = 0$, they both reduce to fully anisotropic and isotropic, respectively. \Citet{heitz2015sggx} provide detailed justification for the use of the SGGX distribution in the context of stochastic microparticle geometry. In our context of stochastic solid geometry, the linear mixture performed better in our ablation study (\cref{tab:projected_full}), so we used it for the rest of our experiments.

\begin{table}
	\centering
	\caption{Chamfer distances on the DTU dataset. \label{tab:dtu_full}}
	\vspace{-1em}
	\begin{tabularx}{\linewidth}{@{} X >{\centering\arraybackslash}X >{\centering\arraybackslash}X >{\centering\arraybackslash}X @{}}
		\toprule
		DTU	& VolSDF	& NeuS	& \textbf{ours} \\
		\midrule
		24	& 2.25	& 3.57	& \textbf{1.92} \\
		37	& 3.19	& 4.02	& \textbf{2.35} \\
		40	& \textbf{1.94}	& 1.99	& 1.96 \\
		55	& 1.61	& 1.71	& \textbf{1.11} \\
		63	& 1.87	& 2.04	& \textbf{1.83} \\
		65	& 2.12	& 2.37	& \textbf{2.01} \\
		69	& 1.61	& 1.70	& \textbf{1.30} \\
		83	& 1.60	& 2.33	& \textbf{1.53} \\
		97	& 2.18	& 2.38	& \textbf{1.62} \\
		105	& 1.73	& 3.17	& \textbf{1.50} \\
		106	& 0.94	& 1.07	& \textbf{0.71} \\
		110	& 1.91	& \textbf{1.09}	& 1.56 \\
		114	& 1.51	& 1.16	& \textbf{0.83} \\
		118	& \textbf{1.36}	& 1.37	& 1.52 \\
		122	& \textbf{1.74}	& 1.83	& 1.78 \\
		\midrule
		mean	& 1.84	& 2.17	& \textbf{1.57} \\
		median	& 1.74	& 1.99	& \textbf{1.56} \\
		\bottomrule
	\end{tabularx}
	\vspace{-1em}
\end{table}

\begin{table}
	\centering
	\caption{Chamfer distances on the NeRF Realistic Synthetic dataset. \label{tab:nerf_full}}
	\vspace{-2em}
	\begin{tabularx}{\linewidth}{@{} X >{\centering\arraybackslash}X >{\centering\arraybackslash}X >{\centering\arraybackslash}X @{}}
		\toprule
		NeRF RS	& VolSDF	& NeuS	& \textbf{ours} \\
		\midrule
		\textsc{chair}	& \textbf{0.0105}	& 0.0182	& 0.0196 \\
		\textsc{lego}	& 0.0354	& 0.0555	& \textbf{0.0156} \\
		\textsc{drums}	& 0.6081	& 0.9897	& \textbf{0.1769} \\
		\textsc{ficus}	& \textbf{0.0279}	& 0.0796	& 0.0436 \\
		\textsc{hotdog}	& 0.1654	& 0.2104	& \textbf{0.1295} \\
		\textsc{materials}	& 0.0231	& 0.0444	& \textbf{0.015} \\
		\textsc{mic}	& 0.9141	& \textbf{0.1214}	& 0.4344 \\
		\textsc{ship}	& 0.2351	& 0.0901	& \textbf{0.0711} \\
		\midrule
		mean & 0.252 & 0.201 & \textbf{0.113} \\
		median & 0.100 & 0.085 & \textbf{0.057} \\
		\bottomrule
	\end{tabularx}
	\vspace{-1em}
\end{table}

\begin{table}
	\centering
	\caption{Chamfer distances on the DTU dataset when using different implicit function distributions $\cdfunc$ for the density $\density$. \label{tab:density_full}}
	\vspace{-1em}
	\begin{tabularx}{\linewidth}{@{} X >{\centering\arraybackslash}X >{\centering\arraybackslash}X >{\centering\arraybackslash}X @{}}
		\toprule
		$\cdfunc$ model	& logistic	& Laplace	& \textbf{Gaussian} \\
		\midrule
		24	& 2.73	& \textbf{1.92} & 1.99 \\
		37	& 3.56	& 3.65	& \textbf{3.08} \\
		40	& \textbf{1.94} & 2.32	& 2.28 \\
		55	& 1.77	& 1.65	& \textbf{1.64} \\
		63	& 1.86	& 1.76	& 1.76 \\
		65	& 2.67	& 2.60	& \textbf{2.45} \\
		69	& 1.73	& 1.58	& \textbf{1.31} \\
		83	& 1.85	& 1.94	& \textbf{1.69} \\
		97	& \textbf{1.82} & 2.13	& 1.83 \\
		105	& 1.90	& 2.04	& \textbf{1.74} \\
		106	& 1.09	& 0.98	& 0.98 \\
		110	& 1.98	& 1.92	& \textbf{1.76} \\
		114	& 1.29	& 1.43	& \textbf{0.96} \\
		118	& \textbf{1.39} & 1.54	& 1.67 \\
		122	& 2.11	& 1.89	& \textbf{1.59} \\
		\bottomrule
		mean	& 1.98	& 1.96	& \textbf{1.78} \\
		median	& 1.86	& 1.92	& \textbf{1.74} \\
		\bottomrule
	\end{tabularx}
	\vspace{-1em}
\end{table}

\begin{table}
	\centering
	\caption{Chamfer distances on the DTU dataset when using different distributions of normals $\ndf$ for the projected area $\projected$. \label{tab:projected_full}}
	\vspace{-1em}
	\begin{tabularx}{\linewidth}{@{} X >{\centering\arraybackslash}X >{\centering\arraybackslash}X >{\centering\arraybackslash}X >{\centering\arraybackslash}X >{\centering\arraybackslash}X >{\centering\arraybackslash}X @{}}
		\toprule
		$\ndf$ model	& delta ($\mathrm{ReLU}$)	& delta	& mixture (const.)	& \textbf{mixture (var.)}	& SGGX (var.) \\
		\midrule
		24	& 3.57	& 2.73	& 2.43	& 2.16	& \textbf{2.10} \\
		37	& 4.02	& 3.56	& 4.16	& 3.40	& \textbf{3.32} \\
		40	& 1.99	& 1.94	& 1.94	& \textbf{1.76}	& 1.83 \\
		55	& 1.71	& 1.77	& 1.85	& \textbf{1.43}	& 1.64 \\
		63	& 2.04	& 1.86	& 1.85	& \textbf{1.60}	& 1.80 \\
		65	& 2.37	& 2.67	& 2.19	& \textbf{1.97}	& 2.34 \\
		69	& 1.70	& 1.73	& 1.57	& \textbf{1.54}	& 1.43 \\
		83	& 2.33	& 1.85	& 1.79	& 1.55	& \textbf{1.49} \\
		97	& 2.38	& \textbf{1.82}	& 2.25	& 1.91	& 2.20 \\
		105	& 3.17	& 1.90	& 1.85	& \textbf{1.53}	& 1.82 \\
		106	& 1.07	& 1.09	& 0.99	& 1.32	& \textbf{0.89} \\
		110	& 1.90	& 1.98	& 1.89	& \textbf{1.59}	& 1.79 \\
		114	& 1.16	& 1.29	& 1.37	& 1.26	& \textbf{1.15} \\
		118	& 1.37	& 1.39	& 1.75	& \textbf{1.31}	& 1.35 \\
		122	& 1.83	& 2.11	& \textbf{1.73}	& 1.85	& 1.95 \\
		\midrule
		mean	& 2.17	& 1.98	& 1.97	& \textbf{1.75}	& 1.81 \\
		median	& 1.99	& 1.86	& 1.85	& \textbf{1.59}	& 1.80 \\
		\bottomrule
	\end{tabularx}
	\vspace{-1em}
\end{table}
\section{Proofs}\label{sec:proofs}

We provide proofs for our theory in \cref{sec:solids,sec:prior}.

\subsection{Exponential transport in opaque solids}\label{sec:proof_theorem}

We begin with the proof of \cref{the:coefficient}. Our proof requires background on continuous-time discrete-space Markov processes, for which we refer to \citet{norris1998markov} as one of many excellent textbooks on this subject matter. We provide pointers to specific parts throughout the proof.

To sketch the proof: 
The free-flight distribution $\freeflight_{\point,\direction}\paren{\distance}$ is the probability density that, starting from $\indicator\paren{\point}=0$, the first $0\to 1$ transition of the indicator function along the ray (i.e., first intersection) happens at distance $\distance$. For this distance to be an exponential random variable, the counting process of $0\to 1$ transitions must be a \emph{Poisson process}. Then, the continuous-time discrete-space random process $\indicator_{\point, \direction}\paren{\distance}$ must be Markovian,%
\footnote{Readers familiar with the derivation of exponential transport for stochastic microparticle geometry will recognize this argument: In that setting, the free-flight distribution describes the distance to the first collision with a microparticle. If microparticle locations are independent, then counting collisions is a Poisson process, and the distance to the first collision is an exponential random variable. In the stochastic solid geometry setting, we replace collisions with $0\to 1$ transitions, and the assumption of independent particle locations with the Markovianity assumption of Equation~\eqref{eqn:markov}.}
and \cref{eqn:coeff_markov} follows from the Kolmogorov equations~\citep{norris1998markov} after enforcing reciprocity and reversibility.

\begin{proof}

We extend Definition~\labelcref{def:general_defs} to define the \emph{conditional transmittance}:
\begin{equation}\label{eqn:ctransmittance_ray}
	\ctransmittance_{\point, \direction}\paren{\distance} \equiv \Prob_{\geometry}\curly{\conditional{\casting_{\point, \direction}\paren{\distance} \ge \distance}{\indicator\paren{\point} = 0}},
\end{equation}
the \emph{conditional free-flight distribution}:
\begin{equation}\label{eqn:cfree_flight}
	\cfreeflight_{\point, \direction}\paren{\distance} \equiv - \frac{\ud \ctransmittance_{\point, \direction}}{\ud \distance}\paren{\distance},
\end{equation}
and the \emph{conditional attenuation coefficient}:
\begin{equation}
	\ccoeff\paren{\point, \direction} \equiv \cfreeflight_{\point, \direction}\paren{0}.\label{eqn:ccoeff}
\end{equation}
Under the assumption of exponential transport, the transmittance equals the conditional transmittance:
\begin{equation}
	\transmittance_{\point, \direction}\paren{\distance} 
	\overset{\mathrm{exponential}}{=} \ctransmittance_{\point, \direction}\paren{\distance}.
\end{equation}
and likewise for the free-flight distribution and attenuation coefficient \citep[Supplementary Section 1, Equation (68)]{bitterli18framework}. Thus, we work with the conditional quantities for this proof.

\paragraph{First jump time.} The restriction  $\indicator_{\point,\direction}$ of the indicator function $\indicator$ along the ray $\ray_{\point, \direction}$ is a \emph{continuous-time discrete-space (binary) stochastic process}. Then, the conditional free-flight distance $\casting_{\point, \direction} \equiv \max\curly{\conditional{\distance\in\leftinc{0,\infty}: \visibility_{\point, \direction}\paren{\distance} = 1}{\indicator_{\point, \direction}\paren{0} = 0}}$ equals the \emph{first jump time} $\jump_{\point, \direction} \equiv \min\curly{\conditional{\distance\in\leftinc{0,\infty}: \indicator_{\point,\direction}\paren{\distance} = 1}{\indicator_{\point, \direction}\paren{0} = 0}}$ of this process; that is, the distance travelled along the ray until the \emph{first} $0 \to 1$ transition of $\indicator_{\point,\direction}$, or equivalently until we first move from outside to inside the opaque solid~\citep[Section 2.2]{norris1998markov}. Consequently, the conditional free-flight distribution $\cfreeflight_{\point, \direction}$ equals the probability density function of $\jump_{\point, \direction}$.

\paragraph{Markov property.} From \cref{eqn:freeflight_exponential}, exponential transport requires that the conditional free-flight distance $\casting_{\point, \direction}$, and thus the first jump time $\jump_{\point, \direction}$, be exponential random variables. This is equivalent to the process $\indicator_{\point,\direction}$ being \emph{Markovian}, that is, satisfying \cref{eqn:markov}~\citep[Section 2.6]{norris1998markov}.

\paragraph{Transition coefficients and generator matrix.} The Markov property implies that we can associate with the process $\indicator_{\point,\direction}$ a \emph{generator matrix}~\citep[Section 2.8]{norris1998markov}:
\begin{equation}
	\genmat_{\point,\direction}\paren{\distance} \equiv \begin{bmatrix}
		-\coeffzo_{\point,\direction}\paren{\distance} & \coeffzo_{\point,\direction}\paren{\distance} \\
		\coeffoz_{\point,\direction}\paren{\distance} & -\coeffoz_{\point,\direction}\paren{\distance}
	\end{bmatrix},
\end{equation}
where $\coeffzo_{\point,\direction}$ and $\coeffoz_{\point,\direction}$ are nonnegative \emph{transition rates}. The transition rate $\coeffzo_{\point,\direction}$ is the rate of the exponential distribution of the first jump time $\jump_{\point,\direction}$, and thus we can relate it to the conditional free-flight distribution as:
\begin{equation}\label{eqn:jump_time}
	\cfreeflight_{\point,\direction}\paren{\distance} = \coeffzo_{\point,\direction}\paren{\distance}\exp\paren{-\int_0^\distance \coeffzo_{\point,\direction}\paren{\distanceVariable} \ud \distanceVariable}. 
\end{equation}
Analogously, the transition rate $\coeffoz_{\point,\direction}$ is the rate of the exponential distribution of the first jump time for $1 \to 0$ transitions (moving from inside to outside the opaque solid). Comparing \cref{eqn:jump_time,eqn:freeflight_exponential}, we see that the transition rate $\coeffzo_{\point,\direction}$ is the restriction of the attenuation coefficient $\coeff$ along the ray $\ray_{\point,\direction}$. Even though we need only $\coeffzo_{\point,\direction}$ and not $\coeffoz_{\point,\direction}$ for exponential transport, our proof characterizes both.

\paragraph{Kolmogorov equations.} To proceed, we will use the \emph{transition probabilities} of the process $\indicator_{\point, \direction}$ for all $i,j \in \curly{0,1}$:
\begin{equation}
	\probij_{\point, \direction}\paren{\distance} \equiv \Prob\paren{\conditional{\indicator_{\point, \direction}\paren{\distance} = j}{\conditional{\indicator_{\point, \direction}\paren{\distance} = i}}},
\end{equation}
and the corresponding \emph{transition probability matrix}:
\begin{equation}
	\transmat_{\point, \direction}\paren{\distance} \equiv \begin{bmatrix}
		\probzz_{\point, \direction}\paren{\distance} & \probzo_{\point, \direction}\paren{\distance} \\
		\proboz_{\point, \direction}\paren{\distance} & \proboo_{\point, \direction}\paren{\distance} \\
	\end{bmatrix}.
\end{equation}
Because $\indicator_{\point, \direction}$ is Markovian, this matrix satisfies the \emph{Kolmogorov forward equation}~\citep[Theorem 2.8.2]{norris1998markov}:
\begin{equation}\label{eqn:kolmogorov_conditional}
	\frac{\ud \transmat_{\point, \direction}}{\ud \distance}\paren{\distance} = \transmat_{\point, \direction}\paren{\distance} \cdot \genmat_{\point, \direction}\paren{\distance}.
\end{equation}

To proceed, we will derive a version of \cref{eqn:kolmogorov_conditional} that uses the unconditional \emph{probability matrix}:
\begin{equation}
	\probmat_{\point, \direction}\paren{\distance} \equiv \begin{bmatrix}
		\Prob\paren{\indicator_{\point, \direction}\paren{\distance} = 0} \\
		\Prob\paren{\indicator_{\point, \direction}\paren{\distance} = 1}
	\end{bmatrix} = \begin{bmatrix}
		\vacancy_{\point, \direction}\paren{\distance} \\
		\occupancy_{\point, \direction}\paren{\distance}
	\end{bmatrix}.
\end{equation}
Using marginalization, we can relate $\probmat_{\point,\direction}$ and $\transmat_{\point,\direction}$ as:
\begin{equation}\label{eqn:uncond_prob_matrix}
	\probmat_{\point,\direction}\paren{\distance} = \transmat_{\point,\direction}\paren{\distance}^\top \cdot \probmat_{\point,\direction}\paren{0}.
\end{equation}
Differentiating with respect to distance, we have:
\begin{equation}
	\frac{\ud \probmat_{\point, \direction}}{\ud \distance}\paren{\distance} = \frac{\ud \transmat_{\point, \direction}}{\ud \distance}\paren{\distance}^\top \cdot \probmat_{\point,\direction}\paren{0}.
\end{equation}
Using the Kolmogorov forward equation \labelcref{eqn:kolmogorov_conditional}:
\begin{equation}
	\frac{\ud \probmat_{\point, \direction}}{\ud \distance}\paren{\distance} = \genmat_{\point, \direction}\paren{\distance}^\top \cdot \transmat_{\point, \direction}\paren{\distance}^\top \cdot \probmat_{\point,\direction}\paren{0}.
\end{equation}
Lastly, using \cref{eqn:uncond_prob_matrix}, we arrive at what we term the \emph{unconditional Kolmogorov equation}:
\begin{equation}\label{eqn:kolmogorov_unconditional}
	\frac{\ud \probmat_{\point, \direction}}{\ud \distance}\paren{\distance} = \genmat_{\point, \direction}\paren{\distance}^\top \cdot \probmat_{\point, \direction}\paren{\distance}.
\end{equation}

\paragraph{Relating ray to local quantities.} An important property of the unconditional Kolmogorov equation \labelcref{eqn:kolmogorov_unconditional} is that it is \emph{local}, that is, it uses quantities defined at only one point, $\ray_{\point,\direction}\paren{\distance}$. In particular, even though our derivation so far has considered a ray starting at point $\point$ and traveling at direction $\direction$, \cref{eqn:kolmogorov_unconditional} requires knowledge of only the direction of the ray not its history (origin $\point$ or distance $\distance$).

We will use this property to rewrite \cref{eqn:kolmogorov_unconditional} in terms of \emph{local} quantities. We associate with every point $\point \in \R^3$ and direction $\direction \in \Sph^2$ the \emph{local} transition rates $\coeffzo\paren{\point, \direction}$ and $\coeffoz\paren{\point, \direction}$, generator matrix
\begin{equation}\label{eqn:generator_local}
	\genmat\paren{\point,\direction} \equiv \begin{bmatrix}
		-\coeffzo\paren{\point,\direction} & \coeffzo\paren{\point,\direction} \\
		\coeffoz\paren{\point,\direction} & -\coeffoz\paren{\point,\direction}
	\end{bmatrix},
\end{equation}
and unconditional probability matrix
\begin{equation}\label{eqn:probability_local}
	\probmat\paren{\point} \equiv \begin{bmatrix}
		\vacancy\paren{\point} \\
		\occupancy\paren{\point}
	\end{bmatrix}.
\end{equation}
It follows that we can relate ray to local quantities as:
\begin{align}\label{eqn:ray_to_local}
	\coeffzo_{\point, \direction}\paren{\distance} &= \coeffzo\paren{\ray_{\point, \direction}\paren{\distance}, \direction}, \\
	\coeffoz_{\point, \direction}\paren{\distance} &= \coeffoz\paren{\ray_{\point, \direction}\paren{\distance}, \direction}, \\
	\genmat_{\point, \direction}\paren{\distance} &= \genmat\paren{\ray_{\point, \direction}\paren{\distance}, \direction}, \\
	\probmat_{\point, \direction}\paren{\distance} &= \probmat\paren{\ray_{\point, \direction}\paren{\distance}}.
\end{align}
From \cref{eqn:ray_to_local,eqn:jump_time,eqn:freeflight_exponential}, we recognize the local transition rate $\coeffzo\paren{\point,\direction}$ as the attenuation coefficient $\coeff\paren{\point,\direction}$.
Additionally, for any quantity $\paren{\ast}$:
\begin{equation}\label{eqn:ray_to_local_deriv}
	\frac{\ud \paren{\ast}_{\point,\direction}}{\ud \distance}\paren{\distance} = \direction \cdot \nabla_{\point} \paren{\ast}\paren{\ray_{\point, \direction}}.
\end{equation}

\paragraph{Local form of the Kolmogorov equation.} Using the correspondences between ray and local quantities in \crefrange{eqn:ray_to_local}{eqn:ray_to_local_deriv}, we can rewrite the unconditional Kolmogorov equation \cref{eqn:kolmogorov_unconditional} in \emph{local form}:
\begin{equation}\label{eqn:kolmogorov_interm}
	\direction \cdot \nabla_{\point} \probmat\paren{\point} = \genmat\paren{\point, \direction}^\top \cdot \probmat\paren{\point}.
\end{equation}
Expanding $\genmat$ and $\probmat$ from \cref{eqn:generator_local,eqn:probability_local}, and using $\occupancy\paren{\point} = 1 - \vacancy\paren{\point}$, we arrive at a scalar equation that we term the \emph{local form of the unconditional Kolmogorov equation}:
\begin{align}
	&\direction \cdot \nabla \vacancy\paren{\point} = \nonumber \\ 
	&\qquad - \vacancy\paren{\point} \coeffzo\paren{\point, \direction} + \paren{1 - \vacancy\paren{\point}} \coeffoz\paren{\point, \direction}.\label{eqn:kolmogorov_unconditional_local}
\end{align}
Exponential transport requires that \cref{eqn:kolmogorov_unconditional_local} hold for rays passing through point $\point$ along any direction $\direction$. In particular, if we consider a ray in direction $-\direction$, we arrive at:
\begin{align}
	&\!\!- \direction \cdot \nabla \vacancy\paren{\point} = \nonumber \\
	&\!\!\qquad - \vacancy\paren{\point} \coeffzo\paren{\point, -\direction} + \paren{1 - \vacancy\paren{\point}} \coeffoz\paren{\point, -\direction}.\label{eqn:kolmogorov_unconditional_local_opposite}
\end{align}

\paragraph{Enforcing reciprocity.} At any point $\point$ and direction $\direction$, \cref{eqn:kolmogorov_unconditional_local,eqn:kolmogorov_unconditional_local_opposite} provide two linear constraints on the four transition rates $\coeffzo\paren{\point, \direction}, \coeffzo\paren{\point, -\direction}, \coeffoz\paren{\point, \direction}, \coeffoz\paren{\point, -\direction}$ as functions of the vacancy $\vacancy\paren{\direction}$ and its directional derivative. Enforcing reciprocity provides an additional linear constraint through the requirement of \cref{eqn:reciprocity_sigma}:
\begin{equation}\label{eqn:reciprocity_transition}
	\coeffzo\paren{\point, \direction} = \coeffzo\paren{\point, -\direction}.
\end{equation}
\Cref{eqn:kolmogorov_unconditional_local,eqn:kolmogorov_unconditional_local_opposite,eqn:reciprocity_transition} form an \emph{underdetermined} linear system on the four transition rates that, after requiring nonnegativity for the transition rates, admits the family of solutions:
\begin{align}\label{eqn:full_solution}
	&\!\!\!\!\begin{bmatrix}
		\coeffzo\paren{\point, \direction} \\
		\coeffzo\paren{\point, -\direction} \\
		\coeffoz\paren{\point, \direction} \\
		\coeffoz\paren{\point, -\direction}
	\end{bmatrix} = \nonumber \\
	&\!\!\!\!\quad \left\{\!\begin{array}{@{}l@{\,}l@{}}
		\begin{bmatrix}
			\frac{\direction\cdot\nabla \vacancy\paren{\point} + \freevar\paren{\point,\direction} \paren{1 - \vacancy\paren{\point}}}{\vacancy\paren{\point}} \\
			\frac{\direction\cdot\nabla \vacancy\paren{\point} + \freevar\paren{\point,\direction} \paren{1 - \vacancy\paren{\point}}}{\vacancy\paren{\point}} \\
			\frac{2\direction\cdot\nabla \vacancy\paren{\point} + \freevar\paren{\point,\direction} \paren{1 - \vacancy\paren{\point}}}{1 - \vacancy\paren{\point}} \\
			\freevar\paren{\point,\direction}
		\end{bmatrix}\!\!, & \text{if } \direction\cdot\nabla \vacancy\paren{\point} \ge 0, \\
		\begin{bmatrix}
			\frac{-\direction\cdot\nabla \vacancy\paren{\point} + \freevar\paren{\point,\direction} \paren{1 - \vacancy\paren{\point}}}{\vacancy\paren{\point}} \\
			\frac{-\direction\cdot\nabla \vacancy\paren{\point} + \freevar\paren{\point,\direction} \paren{1 - \vacancy\paren{\point}}}{\vacancy\paren{\point}} \\
			\freevar\paren{\point,\direction} \\
			\frac{-2\direction\cdot\nabla \vacancy\paren{\point} + \freevar\paren{\point,\direction} \paren{1 - \vacancy\paren{\point}}}{1 - \vacancy\paren{\point}}
		\end{bmatrix}\!\!, & \text{if }  \direction\cdot\nabla \vacancy\paren{\point} < 0,
	\end{array}\right.
\end{align}
where $\freevar\paren{\point,\direction} = \freevar\paren{\point,-\direction} \ge 0$ is a free variable. We can verify this solution by computing the range and null space of the linear system of \cref{eqn:kolmogorov_unconditional_local,eqn:kolmogorov_unconditional_local_opposite,eqn:reciprocity_transition}. Focusing on the attenuation coefficient $\coeff \equiv \coeffzo$, we can write succinctly:
\begin{align}
	&\coeff\paren{\point,\direction} = \coeff\paren{\point,-\direction} = \nonumber \\
	&\qquad \frac{\abs{\direction\cdot\nabla \vacancy\paren{\point}} + \freevar\paren{\point,\direction} \paren{1 - \vacancy\paren{\point}}}{\vacancy\paren{\point}}.\label{eqn:coeff_freevar}
\end{align}

\paragraph{Enforcing reversibility.} \Cref{eqn:coeff_freevar} includes the free variable $\freevar$ whose value we cannot resolve by enforcing only exponential transport (i.e., exponential free-flight distributions) and reciprocity. We can determine the value of $\freevar$ by considering the following: For any two points $\point$ and $\pointalt$, \cref{eqn:kolmogorov_unconditional_local} allows computing the vacancy $\vacancy$ of one point from that of the other, by integrating along the linear segment connecting the two points. As the two vacancies correspond to probabilities of the same underlying random field $\indicator$, we should arrive at consistent results whether we integrate in the direction from $\point$ towards $\pointalt$, or in the reverse direction from $\pointalt$ towards $\point$. Concretely, using the ray notation: 
\begin{tight_enumerate}
	\item In one direction, we start at $\ray_{\point,\point\to\pointalt}\paren{0}=\point$ with initial condition $\vacancy_{\point,\point\to\pointalt}\paren{0} = \vacancy\paren{\point}$, and use \cref{eqn:kolmogorov_unconditional_local} to integrate $\vacancy_{\point,\point\to\pointalt}\paren{\distance}$ with respect to $\distance$ along the ray until we reach $\ray_{\point,\point\to\pointalt}\paren{\norm{\pointalt-\point}}=\pointalt$. The integration result should be $\vacancy_{\point,\point\to\pointalt}\paren{\norm{\pointalt-\point}} = \vacancy\paren{\pointalt}$.
	\item In the reverse direction, we start at $\ray_{\pointalt,\pointalt\to\point}\paren{0}=\pointalt$ with initial condition $\vacancy_{\pointalt,\pointalt\to\point}\paren{0} = \vacancy\paren{\pointalt}$, and use \cref{eqn:kolmogorov_unconditional_local} to integrate $\vacancy_{\pointalt,\pointalt\to\point}\paren{\distance}$ with respect to $\distance$ along the ray until we reach $\ray_{\pointalt,\pointalt\to\point}\paren{\norm{\pointalt-\point}}=\point$. The integration result should be $\vacancy_{\pointalt,\pointalt\to\point}\paren{\norm{\pointalt-\point}} = \vacancy\paren{\point}$.
\end{tight_enumerate}
We term this property \emph{reversibility},%
\footnote{This property is different from the usual notion of time reversibility in continuous-time Markov processes \citep[Section 3.7]{norris1998markov}.}
as it corresponds to enforcing consistency between the process $\indicator_{\point,\direction}\paren{\distance}$ and its directional reverse $\indicator_{\point,-\direction}\paren{\distance}$. Reversibility along all rays is necessary for the underlying indicator function $\indicator$ to represent a solid, as otherwise views of this solid from different rays would have inconsistent geometry. From \cref{eqn:kolmogorov_unconditional_local}, reversibility holds only if the right-hand side of this equation is proportional to the vacancy $\vacancy\paren{\point}$. In turn, from \cref{eqn:full_solution}, this is only true if $\freevar\paren{\point,\direction} \propto \nicefrac{\vacancy\paren{\point}}{1-\vacancy\paren{\point}}$. Setting $\freevar\paren{\point,\direction} \coloneqq 0$ produces the \emph{minimal} transition rates that guarantee reversibility. Then, \cref{eqn:coeff_freevar} simplifies to:
\begin{equation}
	\coeff\paren{\point,\direction} = \coeff\paren{\point,-\direction} = 
	\frac{\abs{\direction\cdot\nabla \vacancy\paren{\point}}}{\vacancy\paren{\point}}.
\end{equation}
This concludes the proof.
 
\end{proof}

\subsection{Stochastic implicit geometry}\label{sec:proof_proposition}

We now prove \cref{pro:implicit}.

\begin{proof}
\Cref{eqn:vacancy_implicit} follows from the definitions of the vacancy function and the indicator function relative to the implicit function in Definition~\ref{def:implicit}:
\begin{align}
\vacancy\paren{\point} &= \Prob\paren{\indicator\paren{\point} = 0} \\
                    &= \Prob\paren{\implicit\paren{\point} > 0} \\
                    &= 1 - \cdf_{\implicit\paren{\point}}\paren{0} \\
                    &= 1 - \cdfunc\paren{-s\meanimplicit\paren{\point}} \\
                    &= \cdfunc\paren{s\meanimplicit\paren{\point}},
\end{align}
where in the last step, we used the assumption that $\cdfunc$ is symmetric. \Cref{eqn:occupancy_implicit} is a consequence of the property $\occupancy\paren{\point} = 1 - \vacancy\paren{\point}$. Lastly, \cref{eqn:coeff_implicit} follows by combining \cref{eqn:vacancy_implicit,eqn:density}, and the property that the PDF $\pdfunc$ is the derivative of the CDF $\cdfunc$. This concludes the proof.

\end{proof}

We close this subsection by adapting \cref{pro:implicit} to the case of a spatially varying scale $s\paren{\point}$. This case is required for adaptive shell representations \citep{zian2023adaptive} and stochastic Poisson surface reconstruction \citep{sellan2022stochastic} (\cref{sec:pointcloud}). \Cref{eqn:occupancy_implicit,eqn:vacancy_implicit} remain unchanged: 
\begin{align}
	\occupancy\paren{\point} &= \cdfunc\paren{-s\paren{\point}\meanimplicit\paren{\point}},\label{eqn:occupancy_implicit_adaptive}\\
	\vacancy\paren{\point} &= \cdfunc\paren{s\paren{\point}\meanimplicit\paren{\point}}. \label{eqn:vacancy_implicit_adaptive}
\end{align}
Combining \cref{eqn:vacancy_implicit_adaptive,eqn:density} results in:
\begin{equation}
	\!\!\!\!\!\mymathboxnt{intTealLL}{\density\paren{\point} \!=\! \frac{\pdfunc\paren{s\paren{\point}\meanimplicit\paren{\point}}\norm{s\paren{\point}\nabla \meanimplicit\paren{\point}\!+\!\nabla s\paren{\point} \meanimplicit\paren{\point}}}{\cdfunc\paren{s\paren{\point}\meanimplicit\paren{\point}}}}. \label{eqn:density_implicit_adaptive}
\end{equation}

\subsection{Density for logistic distribution}\label{sec:proof_logistic}

Lastly, we prove \cref{eqn:coeff_implicit_logistic}. This equation follows easily from \cref{eqn:coeff_implicit} using the properties of the logistic CDF:
\begin{align}
\pdfunc_{\mathrm{logistic}}\paren{s} &\equiv \frac{\ud \cdfunc_{\mathrm{logistic}}}{\ud s}\paren{s} \\
&= \cdfunc_{\mathrm{logistic}}\paren{s} \cdot \paren{1 - \cdfunc_{\mathrm{logistic}}\paren{s}} \\
&= \cdfunc_{\mathrm{logistic}}\paren{s} \cdot \cdfunc_{\mathrm{logistic}}\paren{-s}.
\end{align}

\end{document}